\documentclass{article}

 \usepackage[nonatbib,preprint]{neurips_2021}

\usepackage{graphicx}

\usepackage[utf8]{inputenc} 
\usepackage[T1]{fontenc}    
\usepackage{hyperref}
\usepackage{url}            
\usepackage{booktabs}       
\usepackage{amsfonts}       
\usepackage{nicefrac}       
\usepackage{microtype}      
\usepackage{xcolor}         

\usepackage{times}
\usepackage{epsfig}

\usepackage{amsmath}
\usepackage{amssymb}
\usepackage{subfigure}
\usepackage{caption}
\usepackage{multirow}
\usepackage{algorithm}
\usepackage{algpseudocode}

\usepackage{mathtools}

\title{Hindsight Goal Ranking on Replay Buffer \\for Sparse Reward Environment}

\author{%
	Tung M. Luu \\
	Department of Electrical Electronic\\
	KAIST\\
	Daejeon, Korea \\
	\texttt{tungluu@kaist.ac.kr} \\
	 \And
	Chang D. Yoo \\
	Department of Electrical Electronic\\
	KAIST\\
	Daejeon, Korea \\
	\texttt{cd\_yoo@kaist.ac.kr} \\
}

\begin{document}
	
	\maketitle
	
	\begin{abstract}
		This paper proposes a method for prioritizing the replay experience referred to as Hindsight Goal Ranking (HGR) in overcoming the limitation of Hindsight Experience Replay (HER) that generates hindsight goals based on uniform sampling. HGR samples with higher probability on the states visited in an episode with larger temporal difference (TD) error, which is considered as a proxy measure of the amount which the RL agent can learn from an experience. The actual sampling for large TD error is performed in two steps: first, an episode is sampled from the relay buffer according to the average TD error of its experiences, and then, for the sampled episode, the hindsight goal leading to larger TD error is sampled with higher probability from future visited states. The proposed method combined with Deep Deterministic Policy Gradient (DDPG), an off-policy model-free actor-critic algorithm, accelerates learning significantly faster than that without any prioritization on four challenging simulated robotic manipulation tasks. The empirical results show that HGR uses samples more efficiently than previous methods across all tasks.  
	\end{abstract}
	
	\section{Introduction} \label{sec:introduction}
	
	Reinforcement Learning (RL) \cite{sutton2018reinforcement} is receiving escalating attention as a result of its successes in high profile tasks that include exceeding human-level performance in playing video games \cite{mnih2013playing, mnih2015human}, defeating a Go master \cite{silver2016mastering, silver2017mastering}, and learning to accomplish simple robotic tasks autonomously \cite{ng2006autonomous, kim2004autonomous, levine2016end, lillicrap2015continuous, duan2016benchmarking, kalashnikov2018qt}. 
	
	Despite the many accomplishments, considerable challenges lie ahead in transferring these successes to the complex real-world tasks. An important challenge that must be addressed is to design a more sample efficient reinforcement learning algorithm, especially in sparse reward environments. To address this issue, Lillicrap et al. propose the Deep Deterministic Policy Gradient (DDPG) \cite{lillicrap2015continuous}, which considers an agent that is capable of learning continuous control such as robot manipulation, navigation, and locomotion. Schaul et al. \cite{schaul2015universal} develop the Universal Value Function Approximators (UVFAs), which allows the value function to generalize over both the states and goals (multi-goal). Moreover, to make the agent learn faster in sparse reward environment, Andrychowicz et al. \cite{andrychowicz2017hindsight} introduce Hindsight Experience Replay (HER) that enables the agent to learn even from undesired outcomes. HER combined with DDPG lets the agent learn to accomplish more complex robotic tasks in sparse reward environment.
	
	In HER, the failed episode is uniformly sampled from replay buffer; subsequently, their goals are also sampled uniformly from any one of the visited states such that the failed episode is transmuted into a successful episode. As a consequence, HER does not consider which visited states and episodes might be most valuable for learning, which is a probable cause for sample inefficiency. It would be more sample efficient if the episodes and goals are prioritized according to their importance. The challenge now is to determine a criterion for measuring the importance of goals for replaying. A recent approach referred to as Energy-Based Prioritization (EBP) \cite{zhao2018energy} proposes an energy-based criterion to measure the significance of an episode. Yet, the goals are sampled uniformly from future visited states within an episode. Even in \cite{zhao2018energy}, the extension of Prioritization Experience Replay (PER) \cite{schaul2015prioritized} samples goals uniformly within an episode. Zhao et al. \cite{zhao2019maximum} introduces a prioritization method such that the policy is encouraged to visit diverse goals. However, this method still treats all goals equally. In this work, the significance of a goal is judged by the Temporal Difference (TD) error, which is an implicit way to measure learning progress \cite{schaul2015prioritized, andre1998generalized}. Within an episode, a future visited state with high TD error will be labeled as hindsight goal more frequently. The episode's significance is measured by the average TD error of the experience with its hindsight goal set as one of the visited states in the episode. 
	
	To summarize, our paper makes the following contributions: We present HGR, a method that prioritizes the experience for choosing the hindsight goal. The proposed method is applicable to any robotic manipulation task that an off-policy multi-goal RL algorithm can be applied. We demonstrated the effectiveness of proposed method on four challenging robotic manipulation tasks. We also compare the sample efficiency of our method with baselines including Vanilla HER \cite{andrychowicz2017hindsight}, Energy-Based Prioritization (EBP) \cite{zhao2018energy}, Maximum Entropy-Regularized Prioritization (MEP) \cite{zhao2019maximum}, and one-step prioritization experience \cite{zhao2018energy}. The empirical result shows that the proposed method converges significantly faster than all baselines. Specifically, two-step ranking uses 2.9 factor less number of samples than vanilla HER, 1.9 times less than one-step prioritization experience, 1.3 times less than EBP, and 1.8 times less than MEP. We also conduct an ablation study experiment to investigate the effect of each step.
	
	The remainder of this paper is organized as follows. Section \ref{section:2} describes the background to solve the problem and related RL algorithms for verification. Section \ref{section:3} summarizes the related works. Section \ref{section:4} presents our proposed method for ranking hindsight goals in the replay buffer. Section \ref{section:5} shows the results of our method and comparison with modern methods. Finally, Section \ref{section:6} concludes this work and discuss the advantages and disadvantages.
	
	\section{Background} \label{section:2}
	In this section, the classical framework of RL and RL's extension to multi-goal setting are introduced, and the two most relevant algorithms to the proposed - Deep Deterministic Policy Gradient and Hindsight Experience Replay - are reviewed. 
	
	\subsection{Reinforcement Learning}
	
	Reinforcement learning algorithm attempts to find the optimal policy for interacting with the unknown environment to maximize the cumulative discounted reward that the agent receives per action performed. It should be noted that the environment will be assumed fully observable. This problem is typically modeled as a Markov Decision Process (MDP). The MDP is composed by a tuple $<\mathcal{S}, \mathcal{A}, R, \mathcal{T}, p_0, \gamma, H>$, where $\mathcal{S}$ is a set of states, $\mathcal{A}$ is a set of actions, $R$ is a reward function, $\mathcal{T}$ is a set of transition probabilities (usually unknown) mapping from current states and actions to future states: $\mathcal{S} \times \mathcal{A} \rightarrow \mathcal{S}$, $\gamma \in [0, 1)$ is a discount factor, $p_0$ is the distribution of initial state, and $ H $ is the horizon of an episode. Here, the horizon is assumed finite. A policy maps the state to the action, $\pi: \mathcal{S} \rightarrow \mathcal{A}$, where the policy can be stochastic or deterministic.
	
	\noindent
	Every episode starts with an initial state $ s_0 $ which is sampled from distribution $ p_0 $. Let the agent be in state $s_t$ at time $t$. Assuming it takes action $a_t \sim \pi(s_t)$ and immediately receives a reward $r_t = R(s_t, a_t)$ from the environment. The environment responses to agent's action and presents new state $ s_{t+1} $ to the agent. Here, the new state $ s_{t+1}$ is sampled from $ \mathcal{T}(.|s_t, a_t)$. The return from a state is defined as a discounted sum of future rewards $R_t = \sum_{i=t}^{H}\gamma^{i-t}r_i(s_i, a_i)$. Here, the return depends on the chosen actions, and therefore on the policy $ \pi $, and may be stochastic. The goal of reinforcement learning is to learn a policy which maximizes the expected return from start distribution $\mathbb{E}_{s_0}[R_0|s_0]$. 
	
	The state-action value function is used in many reinforcement learning algorithms to indicate the expected return when an agent is in state $ s_t=s $ and performs action $ a_t=a $ and thereafter follows policy $ \pi $:
	\begin{align}
	Q^{\pi}(s_t, a_t) = \mathbb{E}_{r_{i \geq t}, s_{i > t} \sim \mathcal{T}, a_{i > t} \sim \pi}[R_t|s_t, a_t]
	\end{align}
	Many approaches in reinforcement learning make use of the well-known recursive relationship Bellman equation:
	\begin{equation}
	\begin{aligned}
	Q^{\pi}\left(s_t, a_t\right) &= \\
	R(s_{t}, a_{t}) &+\gamma\mathbb{E}_{s_{t+1} \sim \mathcal{T}}\left[\mathbb{E}_{a_{t+1} \sim \pi}\left[Q^{\pi}\left(s_{t+1}, a_{t+1}\right)\right]\right]
	\end{aligned}
	\end{equation}
	If the policy is deterministic we can denote it as a function $ \mu: \mathcal{S} \rightarrow \mathcal{A} $ and the inner expectation is avoided:
	\begin{align}
	Q^{\mu}\left(s_{t}, a_{t}\right)= R(s_{t}, a_{t})+\gamma\mathbb{E}_{s_{t+1} \sim \mathcal{T}}\left[Q^{\mu}\left(s_{t+1}, \mu(s_{t+1})\right)\right].
	\end{align}
	Let $\pi^*$ denote an \textit{optimal policy}, its state-action value satisfies the condition $Q^{\pi^*}(s, a) \geq Q^{\pi}(s, a)$ for every $s \in \mathcal{S}, a \in \mathcal{A}$, and any policy $\pi$. All optimal policies- there could be multiple- have the same Q value referred to as the \textit{optimal Q-value function}, denoted by $Q^*$. This $Q^*$ satisfies the \textit{optimal Bellman} equation,
	\begin{align}
	Q^*(s_t, a_t) = R(s_t, a_t) + \gamma \mathbb{E}_{s_{t+1} \sim \mathcal{T}}[\max_{a\in \mathcal{A}}Q^*(s_{t+1}, a)]
	\end{align}
	
	\subsection{Deep Deterministic Policy Gradient}
	
\begin{figure}[t]
		\centering
		\includegraphics[width=0.6\textwidth]{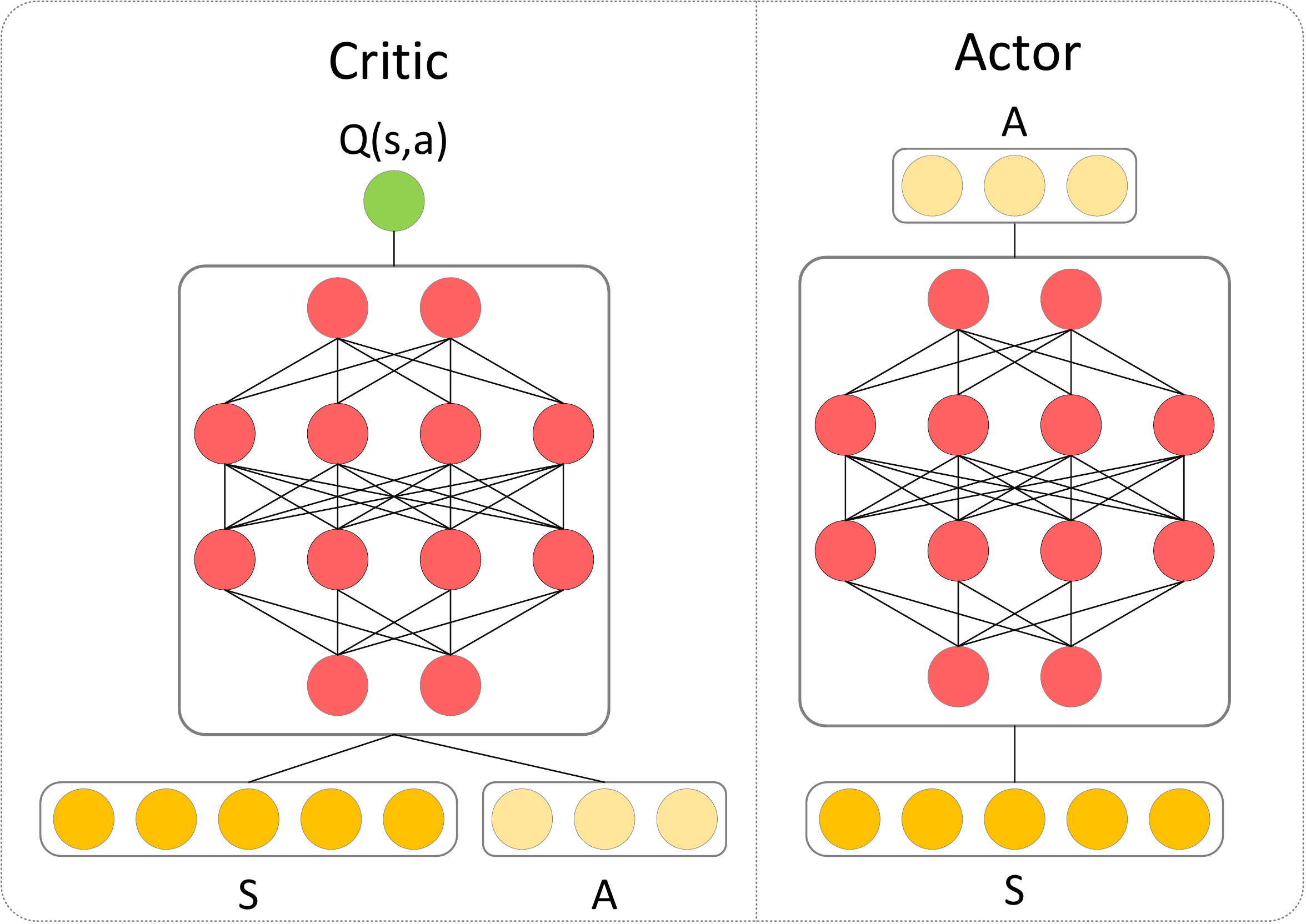}
		\caption{Deep neural network architecture of an actor and a critic in DDPG algorithm.}
		\label{fig_chap_2:2}
	\end{figure}

	\textit{Deep Deterministic Policy Gradient} (DDPG) interleaves learning an approximation to $Q^{\ast}(s,a)$ with learning an approximation to $\mu^{\ast}(s)$, it does so in a way that is specifically adapted for environments with continuous action spaces. Two deep architectures representing critic and actor networks are shown in Figure \ref{fig_chap_2:2}. Here, $Q^{\ast}(s,a)$ is presumed to be differentiable with respect to the action argument. DDPG serves as the backbone of the proposed algorithm, but it need not be the case: we could have used Twin Delayed DDPG (TD3) \cite{fujimoto2018addressing} or Soft Actor-Critic (SAC) \cite{haarnoja2018soft}. 
	
	DDPG uses two deep architectures (for stability, four architectures are used, which is discussed in detail below) in performing actor-critic policy gradient with replay buffer to store real-world experiences to train the actor and critic networks. Figure~\ref{fig_chap_2:2} shows two networks: (1) the actor network that takes the observed state as input and predicts the action that maximizes the action-value function and (2) critic network that takes state and action input pair in predicting the value of the action-value function. The state and action are respectively represented as 5- and 3-dimensional vectors in the figure.
	
	The actor network is trained to maximize the critic network's output for the given state stored in the replay buffer. Here the critic network is assumed well-trained, and its parameters are fixed while updating the actor network. The critic network is trained based on the Bellman optimality equation assuming the actor network is well-trained. The next states of the experiences or tuples stored in the replay buffer and its corresponding predicted action of the actor network, are fed into the critic network in estimating the action-value function for the next state and its predicted action. The critic network is trained such that the action-value function of the current state and its action from the replay buffer matches the one-step look-ahead of the action-value function, which is defined as the sum of the experienced reward stored in the replay buffer and the discounted action-value of the next state and estimated action. 
	
	For obtaining stability during training the critic network, a critic network for training and a critic network for outputting the target value are separately incorporated. Similarly, two separated networks for training and computing next action are also incorporated for learning actor network. The actor-critic networks are repeatedly trained in the manner discussed above. The parameters of the target networks are slowly updated to match the training networks based on moving average. Exploration can be performed by adding noise to the actor network's output or the parameters of the actor network.
	
	Mathematically, DDPG maintains a deterministic policy $\mu_{\theta}(s)$ parameterized by $ \theta $ and a critic $Q_{\phi}(s, a)$ parameterized by $ \phi $. DDPG alternates between running a policy to collect experience and updating the parameters. The episodes are collected by using a behavior policy, which is the version of deterministic policy mixed with noise, i.e., $\pi_b(s) = \mu_{\theta}(s) + \mathcal{N}$, where $\mathcal{N}$ is noise such as mean-zero Gaussian noise or noise generated by Ornstein-Uhlenbeck process. The critic is trained by minimizing the following loss function to encourage the approximated Q-value function to satisfy the Bellman equation:
	\begin{align} \label{bellman_error_loss}
	L(\phi) = \mathbb{E}_{(s, a, r, s') \sim \mathcal{D}} \left[y - Q_{\phi}(s, a)\right]^2 
	\end{align}
	where, $y = r + \gamma Q_{\phi^-}(s', \mu_{\theta^-}(s'))$. Here $ Q_{\phi^-}$, $\mu_{\theta^-} $, and $ \mathcal{D} $ are respectively the target Q-value function, the target policy, and experience replay buffer containing a set of experience $\{(s, a, r, s')\}$.
	
	For stably training, the target $Q_{\phi^-}$ is typically maintained using a separated network, whose weights are periodically updated or averaged over the current weights of main network \cite{mnih2013playing, wang2015dueling, van2016deep}. Subsequently, the actor is learned by maximizing the following objective function with respect to $ \theta $: 
	\begin{align}
	J(\theta) = \mathbb{E}_{s \sim \mathcal{D}} [Q_{\phi}(s, \mu_{\theta}(s))]
	\end{align}
	The derivative of this objective is computed using the deterministic policy gradient theorem \cite{silver2014deterministic},
	\begin{align} \label{dpg_loss}
	\nabla_{\theta}J(\theta) = \mathbb{E}_{s \sim \mathcal{D}} [\nabla_{\theta}\mu_{\theta}(s)\nabla_a Q_{\phi}(s, a)|_{a=\mu_{\theta}(s)}]
	\end{align}
	In this step, the parameter $ \phi $ of Q-value function is kept fixed. To make training more stable, the target policy is similarly maintained by using a separated network, and its weights are updated periodically by moving average.
	
	In practice, the expectation in Eq. (\ref{bellman_error_loss}) and Eq. (\ref{dpg_loss}) is approximated by the mini-batch samples in the experience replay \cite{lin1992self}. Specifically, given a batch of experiences $ B = \{(s_i, a_i, r_i, s'_i)\} $, the mean square error of Bellman equation approximated by,
	\begin{align} \label{bellman_error_loss_minibatch}
	L(\phi) \approx \frac{1}{|B|}\sum_{i=1}^{|B|}\left[r_i + \gamma Q_{\phi^-}(s'_i, \mu_{\theta^-}(s'_i)) - Q_{\phi}(s_i, a_i)\right]^2 
	\end{align}
	and the deterministic policy gradient approximated by,
	\begin{align} \label{dpg_loss_minibatch}
	\nabla_{\theta}J(\theta) \approx \frac{1}{|B|} \sum_{i=1}^{|B|}[\nabla_{\theta}\mu_{\theta}(s_i)\nabla_a Q_{\phi}(s_i, a)|_{a=\mu_{\theta}(s_i)}]
	\end{align}
	
	\subsection{Universal Value Function Approximators}
	
	Universal Value Function Approximators (UVFAs) \cite{schaul2015universal} is an extension of DQN to the multi-goal setting. In this setup, there is more than one goal the agent may try to achieve. This setup is also known as multi-task RL or goal-conditioned RL. Let $ \mathcal{G} $ denote the space of all possible goals. This makes a modification to the reward function $ R $ such that it depends on a state, an action, and a goal $ g \in \mathcal{G} $, i.e. $ r_t = R(s_t, a_t, g) $. For simplicity, we assume the goal space $\mathcal{G}$ is a subset or equal to the state space $\mathcal{S}$, i.e. $\mathcal{G} \subseteq \mathcal{S}$, which is satisfied in our environment. Every episode starts with a initial state and a goal sampled from distribution $ p(s_0, g) $. After sampling, the goal is fixed within the episode. At every timestep, the policy takes as input not only the current state but also the goal of the current episode, $ \mu: \mathcal{S} \times \mathcal{G} \rightarrow \mathcal{A} $. The Q-value function now also depends on the goal $ Q^{\mu}(s_t, a_t, g) = \mathbb{E}_{s_t \sim \mathcal{T}, a_t \sim \mu, g \sim \mathcal{G}}[R_t|s_t, a_t, g] $. In \cite{schaul2015universal}, authors show that it is possible to learn an approximator to Q-value function using direct bootstrapping from the Bellman equation similar to DQN. The extension UVFAs for Deep Deterministic Policy Gradient is straightforward. It results in a modification for updating the critic,
	\begin{align}
	\nabla_{\phi}L(\phi) = \nabla_{\phi}\mathbb{E}_{(s, a, r, s', g) \sim \mathcal{D}}[y - Q_{\phi}(s, a, g)]^2
	\end{align}
	where, $ y = r + \gamma Q_{\phi^-}(s', \mu_{\theta^-}(s', g), g) $. Subsequently, the update for the actor,
	\begin{align}
	\nabla_{\theta}J(\theta) = \mathbb{E}_{(s, g) \sim \mathcal{D}}[\nabla_{\theta}\mu_{\theta}(s, g)\nabla_aQ_{\phi}(s, a, g)|_{a=\mu_{\theta}(s, g)}]
	\end{align}
	
\begin{figure}[t]
		\centering
		\includegraphics[width=0.6\textwidth]{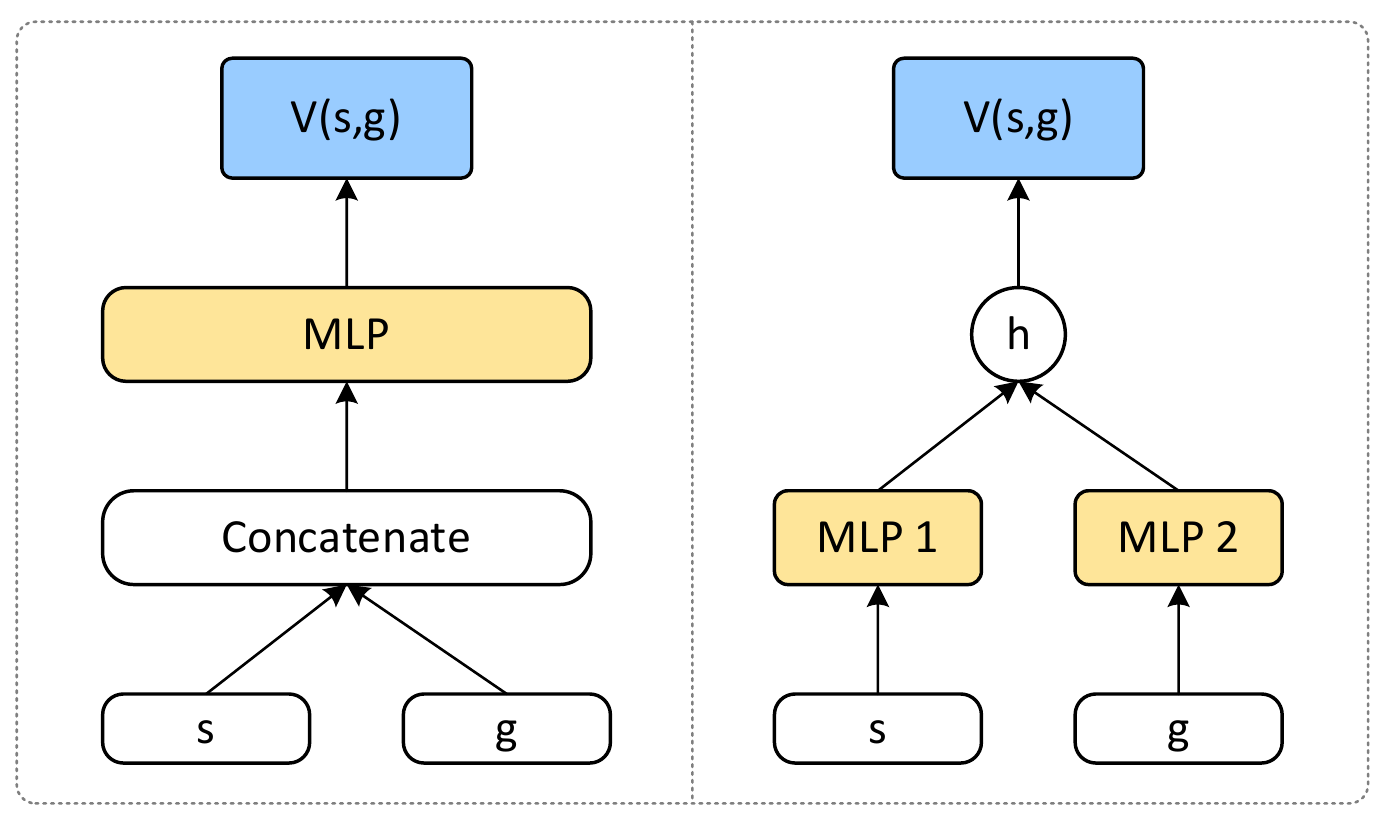}
		\caption{Diagram of the presented function approximation architectures in UVFAs \cite{schaul2015universal}. In blue rectangles, the learning targets for the output of each network (yellow) are shown. \textbf{Left}: concatenated architecture. \textbf{Right}: two-stream architecture with two separate sub-networks MLP-1 and MLP-2 combined at $h$.}
		\label{fig_chap_2:3}
	\end{figure}

	In \cite{schaul2015universal}, a value function approximator that generalizes over both states and goals is considered. Two architectures for approximating the value function are depicted in Figure \ref{fig_chap_2:3}: one architecture directly concatenates state and goal and outputs the value of value function for the concatenated input, and the other architecture processes the states and goals separately before taking their respective outputs in predicting the value of the value function.
	The proposed algorithm is based on the architecture that directly concatenates the state and goal: the architecture is simple and can be easily integrated into various RL algorithms.
	
	\subsection{Hindsight Experience Replay}
	
	Despite a wide range of advances in the application of Deep Reinforcement Learning, learning the agent in an environment with sparse rewards remains a major challenge- especially in robotic tasks where the desired goal is challenging, and the reward is sparse. A reward that is commonly used in a sparse reward environment is as follows:
	
	\begin{align}\label{sparsereward}
	r_t =
	\begin{cases}
	\ \ \ 0,&\text{if $\|s_t-g\|_2 < \rho$}\\
	-1,&\text{otherwise}\\
	\end{cases} 
	\end{align} 
	where, $s_t$, $g$, and $\rho$ are respectively the current state, desired goal, and tunnable hyperparameter. Intuitively, this function rewards an agent if the current state of the agent is close with the desired goal within threshold $\delta$.
	
	In this environment, the agent will not receive any positive reward ``0"  for a long period and will have difficulty in learning. As a result, the agent will be faced with the sample inefficiency problem, which is one of the main concerns in RL. Andrychowicz et al. \cite{andrychowicz2017hindsight} propose a simple yet effective method referred to Hindsight Experience Replay (HER), that relabels the goal of the existing experiences in the replay buffer to overcome the sample inefficiency problem in the sparse reward environment. 
	
	HER duplicates the episodes in replay buffer but with the set of goals such that the episode is either a successful episode or in the future steps of the episode going to be successful. In \cite{andrychowicz2017hindsight}, authors propose four strategies in selecting a visited state for hindsight goal including \texttt{final}, \texttt{future}, \texttt{episode}, and \texttt{random}. In the \texttt{final} strategy, the hindsight goal is the terminal state in the sampled episode. In the \texttt{future} strategy, the hindsight goal is selected randomly from states at future time steps with respected the chosen experience. In the \texttt{episode} strategy, the hindsight goal is an arbitrary state which comes from the same episode of the sampled experience, no matter it is observed after or before a chosen experience. In the \texttt{random} strategy, the hindsight goal is sampled randomly from visited states encountered so far in the whole replay buffer. All strategies are effective, but the \texttt{future} strategy outperforms the other three strategies. In this paper, only the \texttt{future} strategy will be considered, and all experiments will be conducted based on this strategy. HER can integrate with any off-policy RL algorithm assuming that we can find a corresponding goal at any state in state space. This assumption is also satisfied in our environment. 
	
	\section{Related works} \label{section:3}
	
	In sparse reward environments, to fully explore the continuous and high dimensional action space in learning an appropriate policy,
	a naive exploration by adding noise to the action or incorporating $\epsilon$-greedy policy is bounded to fail. For a long horizon, the task becomes exponentially more difficult. Furthermore, in the real world, the number of samples that can be collected in real-world tasks is generally limited, and the sample efficiency becomes critical. Thus, exploring diverse outcomes and learning policies in a sparse reward environment is challenging.
	
	Nair et al. (2018) \cite{nair2018overcoming} approaches this problem using demonstrations combined with HER. Here demonstration is used as a guide for exploration. The proposed algorithm attempts to teach the agent to learn gradually from easy targets to challenging targets, which is a form of curriculum learning. 
	Florensa et al. (2017) \cite{florensa2017reverse}  propose Reverse Curriculum generation. In this method, the agent starts off from an initial state that is right next to the goal state, then two steps away from the goal, and so on. 
	Florensa et al. (2017) \cite{florensa2018automatic} use a generator network to automatically label and propose goals at the appropriate level of difficulty. 
	Forestier et al. (2017) \cite{forestier2017intrinsically} introduce a method based on Intrinsically Motivated Goal Exploration Processes (IMGEPs) architecture that uses learned goal spaces and current learning progress of agent to generate goals with increasing complexity. This method maintains a memory-based representation of policies for performing multi-task. 
	Colas et al. (2018) \cite{colas2019curious} propose a similar method for multi-task multi-goal RL referred to as CURIOUS. The tasks and goals are also selected based on the learning progress. However, CURIOUS uses a single monolithic policy for all tasks and goals rather than memory-based policies.
	
	In the domain of goal-directed RL, Andrychowicz et al. (2017) \cite{andrychowicz2017hindsight} propose a simple yet effective method referred to as Hindsight Experience Replay, which may be considered as a form of implicitly curriculum learning. A number of recent works adapt the idea of HER to different situations. 
	Rauber et al. (2019) \cite{rauber2019hindsight} incorporate HER into Policy Gradient methods. 
	Fang et al. (2019) \cite{fang2018dher} extend HER into a dynamic goal situation, where the desired goal is moving in the episode. 
	Ding et al. (2019) \cite{ding2019goal} attempt to incorporate HER with the well-known Imitation Learning (IL) algorithms such Behavior Cloning (BC) \cite{pomerleau1989alvinn} or Generative Adversarial Imitation Learning (GAIL) \cite{ho2016generative}. 
	
	In HER, the key assumption is that the goals are sampled from a set of states that need to be visited. Nevertheless, the real-world problems like energy-efficient transport, or robotic trajectory tracking, rewards are often complex combinations of desirable rather than sparse objectives. 
	Eysenbach et al. (2020) \cite{eysenbach2020rewriting} propose inverse RL to generalize the goal-relabeling techniques to arbitrary classes of reward functions ranging from multi-task settings, discrete set of rewards, and linear reward functions. In this method, inverse RL is used as relabeling strategy to infer the most likely goal given a trajectory. 
	Similar work is proposed by Li et al. (2020) \cite{li2020generalized}, which uses the approximated inverse RL to sample a suitable goal for a given trajectory.
	Pong et al. \cite{nair2018visual, pong2019skew} generalize state in multi-goal RL into raw pixel and proposes to sample goals from a VAE prior. The group of those methods can be considered as a different strategy to generate hindsight goals.
	
	In this paper, the efficiency of experience replay using replay buffer is studied, and an algorithm is proposed to improve the sample efficiency further. HER is an effective algorithm for reducing sample complexity in a sparse reward environment. However, the uniform sampling from future visited states for hindsight goals that HER is based on could be improved for obtaining better sample efficiency. 
	
	HER replaces the actual goal of an experience with a randomly sampled state visited in the future without considering the significance of the sampled \cite{plappert2018multi}. The recent method referred to as Energy-Based Prioritization (EBP) \cite{zhao2018energy} proposes to prioritize episodes in the replay buffer. Specifically, based on the work-energy principle in physics, authors introduce an episode energy function that measures the importance of an episode. Subsequently, the higher energy episode is replayed more frequently. Episode energy function can be considered an alternative to the TD error in measuring the significance of an episode. However, EBP samples goals uniformly within a sampled episode, which does not consider the importance of future visited states. Moreover, the authors also compare their methods with the extension of Prioritized Experience Replay (PER) \cite{schaul2015prioritized}, which is prioritizing the experiences in replay buffer rather than the episode. However, after the experience is sampled, the goal is still sampled randomly at future time steps with respect to the chosen experience, which is sample inefficiency. Another approach referred to as Maximum Entropy-regularized Prioritization (MEP) \cite{zhao2019maximum} is proposed for encouraging the policy to visit diverse goals. Specifically, the trajectories are prioritized during training the policy such that the distribution of experienced goals in the replay buffer as uniform as possible. However, the goals in a considered episode are still uniformly sampled. In this work, to address the sample inefficiency in HER, we propose a two-step ranking using TD error to judge which goals and episodes are valuable for learning. 
	
	\section{Hindsight Goal Ranking} \label{section:4}
	To learn in the sparse reward environment, Hindsight Experience Replay (HER) relabels the goal of a failed episode by any one of the future visited states such that the episode becomes successful. However, the hindsight goal is sampled uniformly from the future visited states attained in the episode, which may not be the most beneficial way of improving sample efficiency. In this section, we answer the question: ``which hindsight goals should be generated?" and ``how to select the hindsight goals within an episode?". An extension of the HER algorithm is developed based on Deep Deterministic Policy Gradient (DDPG).
	
	\subsection{Two-step prioritized hindsight goal generation}
	
	In HER, the episodes are initially sampled randomly from the experience buffer for replaying. Subsequently, the hindsight goal of a sampled episode is selected by uniformly sampling a future visited state. The replay process does not try to determine which goals and episodes are more valuable for learning \cite{plappert2018multi}. 
	In \cite{zhao2018energy}, Energy-Based Prioritization (EBP) attempts to prioritize an episode in the replay buffer by its energy. However, the future visited states in an episode are sampled uniformly to generate a hindsight goal. Also, in \cite{zhao2018energy}, the extension of Prioritized Experience Replay (PER) tries to prioritize experiences instead. However, within the episode containing chosen experience, the hindsight goal is still selected randomly from a bunch of future visited states. Instead of uniformly sampling future visited states for relabeling to hindsight goals, this paper improves sample efficiency by prioritizing the future visited states within an episode according to the magnitude of the TD error $ \delta $, computed by Eq. (\ref{td_error}). 
	\begin{align}\label{td_error}
	\delta = R(s, a, g) + \gamma Q_{\phi^-}(s',\mu_{\theta^-}(s', g), g) - Q(s, a, g)
	\end{align}
	
	\begin{figure}[t]
		\centering
		\includegraphics[width=0.6\textwidth]{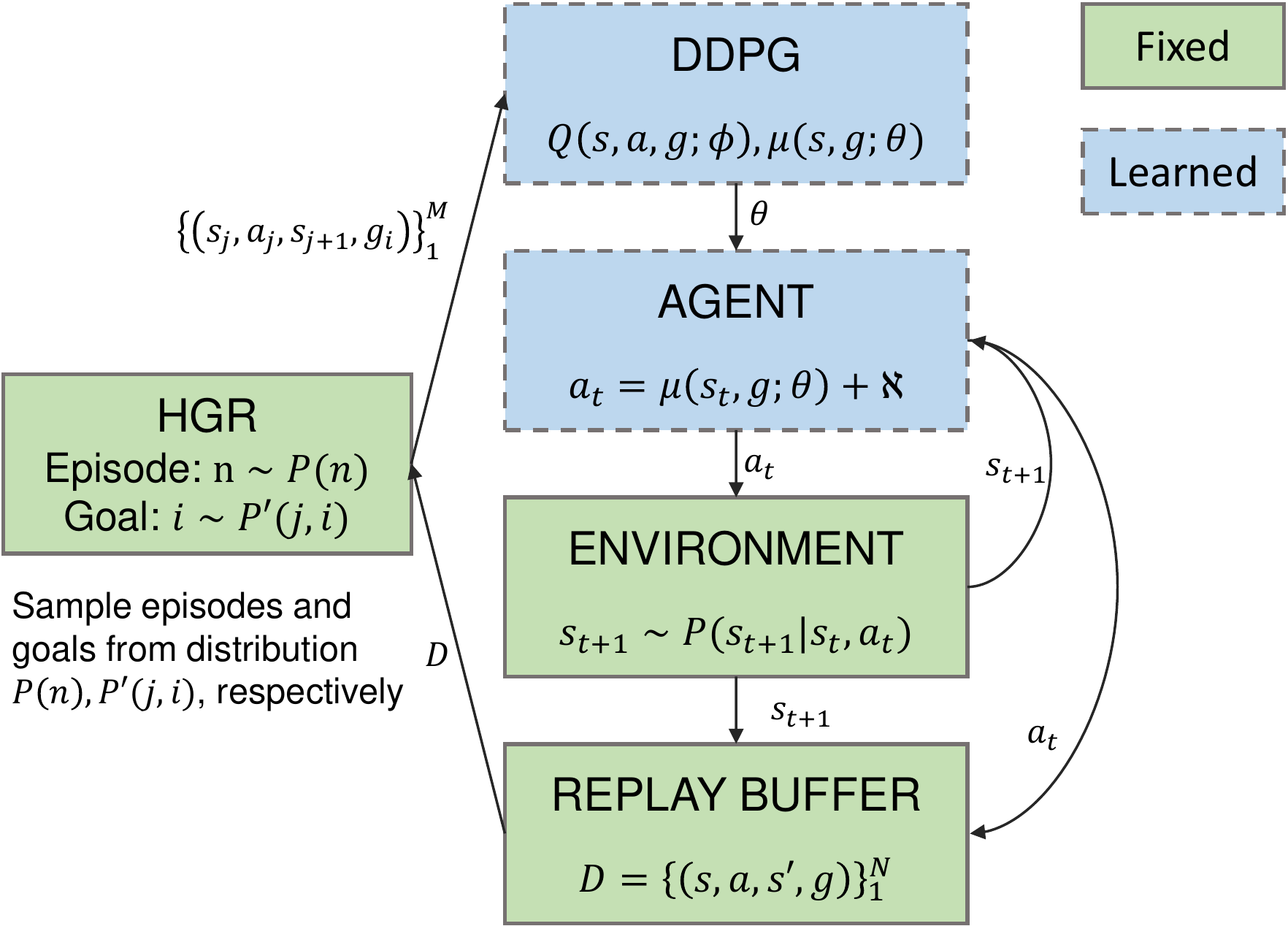}
		\caption{The overview of \textbf{HGR} algorithm: The episode and the experience with its hindsight goal are sampled from distribution (\ref{episodepriority}) and (\ref{transpriority}), respectively. The DDPG and AGENT contains learnable modules.}
		\label{fig_chap_3:1}
	\end{figure}
	
	\noindent
	This criterion has been considered as a proxy measure of the amount which the RL agent can learn from an experience: concretely, the TD error measures how far the value is from its next-step bootstrap estimate \cite{andre1998generalized, schaul2015prioritized}. Using TD error for prioritizing is particularly applicable to DDPG \cite{lillicrap2015continuous}, which needs to compute the TD error to update parameters of Q-value function.
	Specifically, in the replay buffer, the importance of an experience with a future visited state, which is probably become a hindsight goal, is ranked based on the magnitude of its TD error. 
	Note that since we follow the \texttt{future} strategy \cite{andrychowicz2017hindsight}, the future visited state should be ranked relative to the others in a \textit{same} episode. Thus in the considered episode, the future visited state with larger magnitude TD error will be labeled as hindsight goal more frequently such that for each sampled episode, the agent is attempting to maximize what it can learn from an experience. Assuming a sampled episode has $ (H - 1) $ experiences. For the $j^{th}$ experience of the episode, the $i^{th}$ future visited state where $(j+1)\leq i \leq H$ is sampled with the following probability:
	\begin{align} \label{transpriority}
	i\sim P'(j, i) = \frac{1}{Z'}|\delta_{ji}|^{\alpha'}, \ \; i\in \{j+1,\ldots, H\}
	\end{align}
	where the normalization function $ Z' = \sum_{j=1}^{H-1}\sum_{i=j+1}^{H} |\delta_{ji}|^{\alpha'} $, and  $\delta_{ji}$ is the TD error of the $j^{th}$ experience and $i^{th}$ visited state. The exponent $ \alpha' $  is a  hyper-parameter which is preset before sampling. When $\alpha' = 0$, uniform sampling is performed, and the proposed algorithm becomes the vanilla HER. To guarantee that the probability of sampling any of the visited states is non-zero, a small positive constant $ \epsilon $ is added to $\delta_{ji}$.
	
	The agent can achieve better sample efficiency by maximizing what it can learn from the whole replay buffer, which requires prioritizing the episodes as well as the goals. Episodes can be ranked in a similar manner as the goal, where the importance of an episode is measured by the average TD error of all experiences within the episode, i.e., all possible combinations of a chosen experience with a future visited state. Let $ K $ be the number of combinations of experience and goals in an episode. Then, the priority of the $ n^{th} $ episode in the replay buffer is defined as the average TD error by $ K $ such that  $\delta^{(n)} = \frac{1}{K}\sum_k^K |\delta_k^{(n)}|$ where $\delta_k^{(n)}$ is the TD error of the $k^{th}$ experience-goal combination from a total of $ K $ combinations. Finally, the $ n^{th} $ episode is sampled with the probability as
	\begin{align} \label{episodepriority}
	n \sim P(n) = \frac{1}{Z}|\delta^{(n)}|^{\alpha}
	\end{align}
	where the normalization function $ Z = \sum_n |\delta^{(n)}|^{\alpha} $. Here $\alpha$ determines how much  prioritization should be incorporated: $\alpha = 0$ is equivalent to the uniform case (no prioritization), i.e., the episodes have the same probability of being sampled. The small positive number is also added to priority to prevent zero probability.
	
	\begin{algorithm*}[ht]
		\caption{DDPG + Hindsight Goal Ranking}
		\begin{algorithmic}[1]
			\State \textbf{Given}: 
			\begin{itemize}
				\item An off-policy RL algorithm $\mathbb{A}$ \Comment{In our case, $ \mathbb{A} $ is DDPG}
				\item A reward function $r : \mathcal{S} \times \mathcal{A} \times \mathcal{G} \rightarrow \mathbb{R}$
				\item Batch size $M$, step-size $\eta$, empty replay buffer $\mathcal{D}$ size $N$, update frequency $U$, $ L $ epoch training, exponents $\alpha, \beta, \alpha'$, and $\beta'$    
			\end{itemize}            
			\State Initialize neural networks of  $\mathbb{A}$
			\State Initialize replay buffer $\mathcal{D}$
			\For {epoch = 1, $ L $}
			\State Sample a goal $g$ and an initial state $s_0$
			\For{t = 0, $ H $ - 1}
			\Comment{Rollout one episode}
			\State Sample an action $a_t$ using the behavioral policy from $\mathbb{A}$: 
			\State $\quad \quad a_t \leftarrow \mu(s_t\|g) + \mathcal{N}$ \Comment{$\|$ denotes concatenation}
			\State Execute the action $a_t$ and observe a new state $s_{t+1}$
			
			\State Store the transition $(s_t, g, a_t, s_{t+1})$ in $\mathcal{D}$ with maximal priority $\hat{\delta} = \underset{\mathcal{D}}{\max}\delta$
			\EndFor
			\If{episode mod  $U \equiv 0$}
			\For{k = 0 to M-1}
			\State Sample $ n^{th} $ episode for replaying based on $P(n)$ 
			\Comment{episode prioritization}
			\State Sample $(s_j,g_j, a_j, s_{j+1})$ and $g_i$ based on $P'(j, i)$ \Comment{goal prioritization}
			\State $r_j' \leftarrow r(s_j, a_j, g_i)$ \Comment{Recalculate reward (HER)}
			\State Compute importance sampling weight by Eq. (\ref{final_weight})
			\State $\quad \quad w_{ji}^{(n)}=w_{ji}\cdot w_n$
			\State Compute TD error  $\delta_{ji}^{(n)} = r_j' + \gamma_j Q_{\phi^-}(s_{j+1}\|g_i , \mu_{\theta^-}(s_{j+1}\|g_i)) - Q_{\phi}(s_j\|g_{i},a_j)$
			\State Update $P(n) \leftarrow \frac{2}{H(H+1)}\sum_t |\delta_t^{(n)}|$ and $P'(j, i) \leftarrow |\delta_{ji}^{(n)}|$
			\State Accumulate critic's weight  $\Delta \leftarrow \Delta + w_{ji}^{(n)}\cdot \delta_{ji}^{(n)}\cdot \nabla_{\phi}Q_{\phi}(s_j\|g_i,a_j)$
			\State Accumulate actor's weight  $\Delta' \leftarrow \Delta' + \nabla_{\theta}\mu_{\theta}(s_j\| g_i)\nabla_aQ_{\phi}(s_j\|g_i, a)|_{a=\mu_{\theta}(s_j\|g_i)}$
			
			\EndFor
			\State Update critic's weights $\phi \leftarrow \phi + \eta \cdot (\max_{i, j, n} w_{ji}^{(n)})^{-1}\cdot \Delta$ and reset $\Delta = 0$
			\State Update actor's weights $\theta \leftarrow \theta + \eta \cdot \Delta'$ and reset $\Delta' = 0$
			\State Update target network of actor and critic using moving average
			\EndIf
			\EndFor
		\end{algorithmic}
		\label{hgr_algo}
	\end{algorithm*} 
	
	The schematic view of HGR is shown in Figure \ref{fig_chap_3:1}. There are five components including (1) \textbf{DDPG}, which concurrently learns the action-value function $Q$ and a deterministic policy $\mu$, (2) \textbf{AGENT}'s behavior policy which is a mixture of the deterministic policy $ \mu_{\theta} $ and Gaussian noise. The behavior policy takes as input the current state $s_t$ and the pursuing goal $g$, then produces action $a_t$ to interact with the (3) \textbf{ENVIRONMENT}. Subsequently, the environment presents new state $s_{t+1}$ to the agent. Immediately, the action $a_t$ and new state $s_{t+1}$ are stored into (4) \textbf{REPLAY BUFFER} $\mathcal{D}$ with the size of $N$. When learning $Q$ and $\mu$, (5) \textbf{HGR} samples an episode and an experience with its hindsight goal from the distribution (\ref{episodepriority}) and (\ref{transpriority}), respectively. This sampling process is performed $M$ times for obtaining a batch of data for performing DDPG.
	
	\noindent
	\textbf{Implementation:}  
	
	\noindent
	There are a total of $ H(H+1)/2 $ experiences in our experiment. The goals are sampled based on heuristics. The sampling complexity is $ O(n) $, and the complexity for updating a priority is $ O(1) $. The size of the replay buffer is very large (millions), and to sample from the distribution (\ref{episodepriority}), the sum-tree data structure similar to that used in \cite{schaul2015prioritized}, where every parent node is the sum of its children node and the leaf nodes represent the priorities. Its complexity for both sampling and updating is $ O(\log n)$.

	\subsection{Prioritization and Bias trade-off}
	
	The Q-value function is estimated based on stochastic updates of samples drawn from the replay buffer. When samples are drawn uniformly, the estimation is unbiased; however, when the samples are prioritized in the replay buffer, the estimation will be biased as prioritization will not allow samples to be drawn from the distribution that defines expectation the Q-value function. Consequently, the prioritization induces a bias in the estimation of the Q-value function  \cite{schaul2015prioritized} and therefore changes the solution that the estimations will converge to. To overcome this problem, we can correct the bias by using importance-sampling (IS) weights. Specifically, when updating the parameters of approximated Q-value function by using the experiences in sampled episodes, the corresponding gradient is scaled by multiplying with the IS weight of the episode, which is defined as Eq. (\ref{is_episode}). This IS weight helps to scale down the sample's gradient when it is updated frequently and remains the same when it is rarely updated- line 21 in Algorithm \ref{hgr_algo}.
	\begin{align}\label{is_episode}
	w_n = \left(\frac{1}{N_{e}}.\frac{1}{P(n)}\right)^{\beta}
	\end{align}
	where $N_e$ is the number of collected episodes in the replay buffer, $\beta$ is hyper-parameter to control how much bias is corrected. If $\beta=1$, the weight fully compensates for non-uniform probability $P(n)$. 
	In practice, we linearly schedule $\beta$, which starts from $\beta_0$ and ends up with one during training. Here $\beta_0$ is a tunable parameter.
	
	Annealing the bias by multiplying the IS weight of the episode is effective method for reducing the bias. However, the bias induced by the prioritization within an episode still remains. To compensate this bias, a small modification is introduced. As mentioned in the previous section, there are $H(H+1)/2$ combinations of experience-goal pair. Therefore, the IS weight for the $i^{th}$ goal according to the $j^{th}$ experience in an episode is
	\begin{align}\label{is_transition}
	w_{ji} = \left(\frac{2}{H(H+1)}.\frac{1}{P'(j, i)}\right)^{\beta'}
	\end{align}
	where, $H$ is horizon, $\beta'$ plays a role similar with $\beta$ to control bias correction. The final IS weight to correct bias for an experience-goal is computed by
	\begin{align}\label{final_weight}
	w^{(n)}_{ji} = w_n.w_{ji}
	\end{align}
	
	For a new experience with an unknown TD error, the maximal priority,  ($P_t = \max_{i < t}P_i$), is assigned to guarantee that all experiences are replayed at least once. For more stable convergence, the IS weights are normalized by their maximum $ 1/\max w_{ji}^{(n)}$; thus the gradient is always scaled downwards. The detailed algorithm is presented in Algorithm \ref{hgr_algo}. 
	
	HER is based on the assumption the reward function is defined as $ R(s\|g, a) $ which allows the agent to evaluate the reward from any state and goals, in Algorithm \ref{hgr_algo}. This assumption is not very restrictive and can be satisfied in our environment. For example, in the reaching task, the reward function is defined as the Euclidean distance between the current end-effector position and the target position. Therefore, when collecting experience, we need not observe and store reward in the replay buffer- line 9, 10 of the Algorithm \ref{hgr_algo}. We use the epsilon-greedy to alter the uniform random policy and the behavior policy for better exploration. Furthermore, the behavior policy is a mixture of the deterministic policy and Gaussian noise- line 8 of Algorithm. For details regarding noise and $\epsilon$-greedy, see in Section \ref{training_detail}.

	\section{Experiments} \label{section:5}
	
	This section will describe the robot simulation benchmark used for evaluating the proposed method. Then, we will investigate the following questions:
	\begin{itemize}
		\item Does the hindsight goal ranking benefit HER?
		\item Does the hindsight goal ranking improve the sample efficiently in the robot benchmark environment?
		\item What is the benefit of prioritizing the goals versus episodes?
	\end{itemize}	
	
	\subsection{Environmental benchmarking} 
	
	HGR is evaluated on  the 7-DOF Fetch Robotic-arm \footnote{\url{https://gym.openai.com/envs/\#robotics}} simulations environment provided by OpenAI Gym \footnote{\url{https://github.com/openai/gym}} \cite{plappert2018multi, 1606.01540}, using MuJoCo physics engine \cite{mujoco}. The robotic arm environment is based on currently available hardware \footnote{\url{https://fetchrobotics.com/}} and is designed as a standard benchmark for Multi-goal RL \cite{plappert2018multi}. In this environment, the agent is required to complete several manipulation tasks with different objectives in each scenario. A 7-DOF Robotics arm with parallel grippers is used to manipulate an object placed on a table in front of the robot, as shown in Figure \ref{fig_chap_4:1}. There are four different tasks of the robotic arm with the difficulty level increasing:
	
	\begin{figure}[t]
		\centering
		\begin{tabular}{cccc}
			\includegraphics[width=0.22\textwidth]{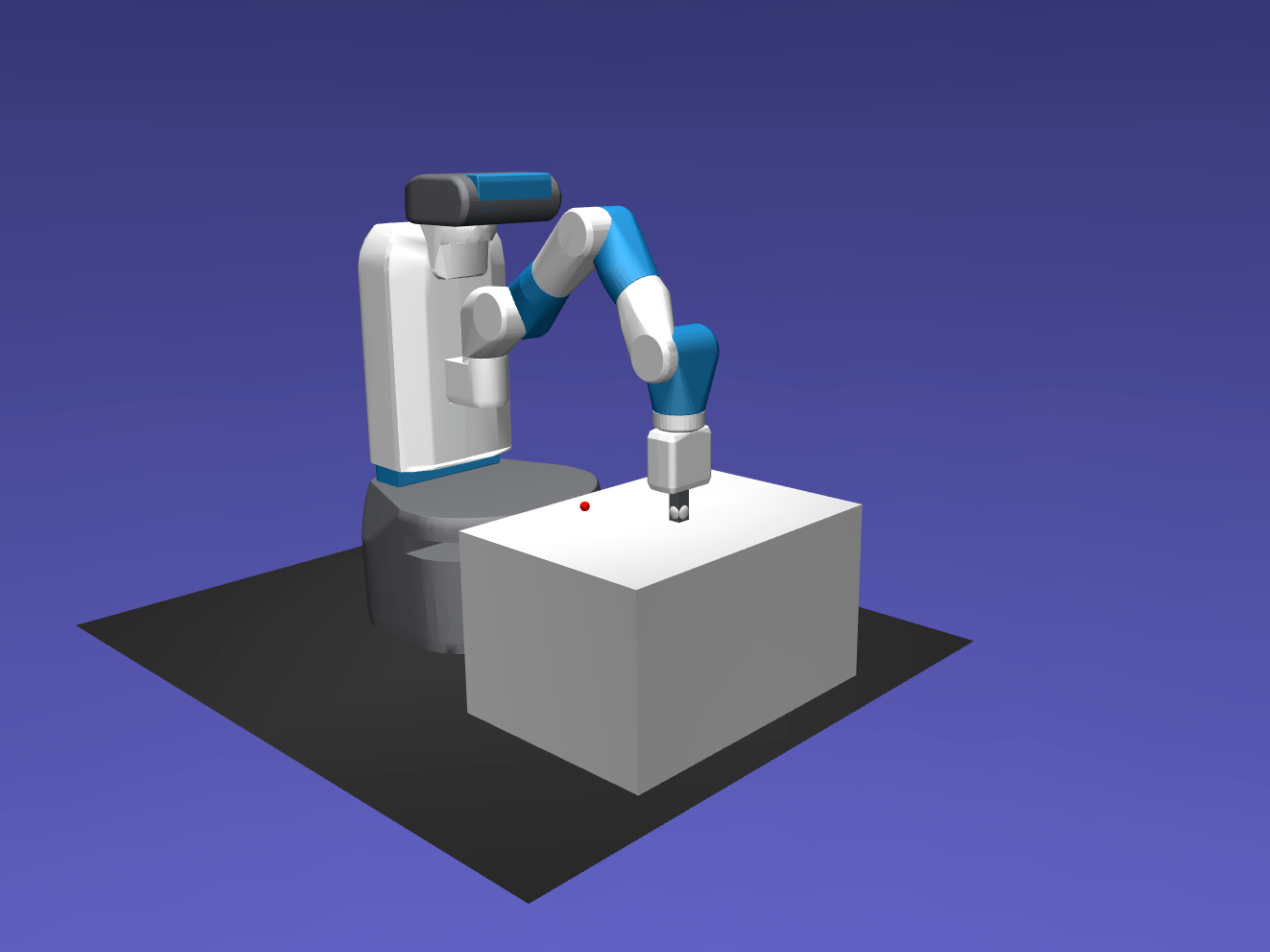} &   \includegraphics[width=0.22\textwidth]{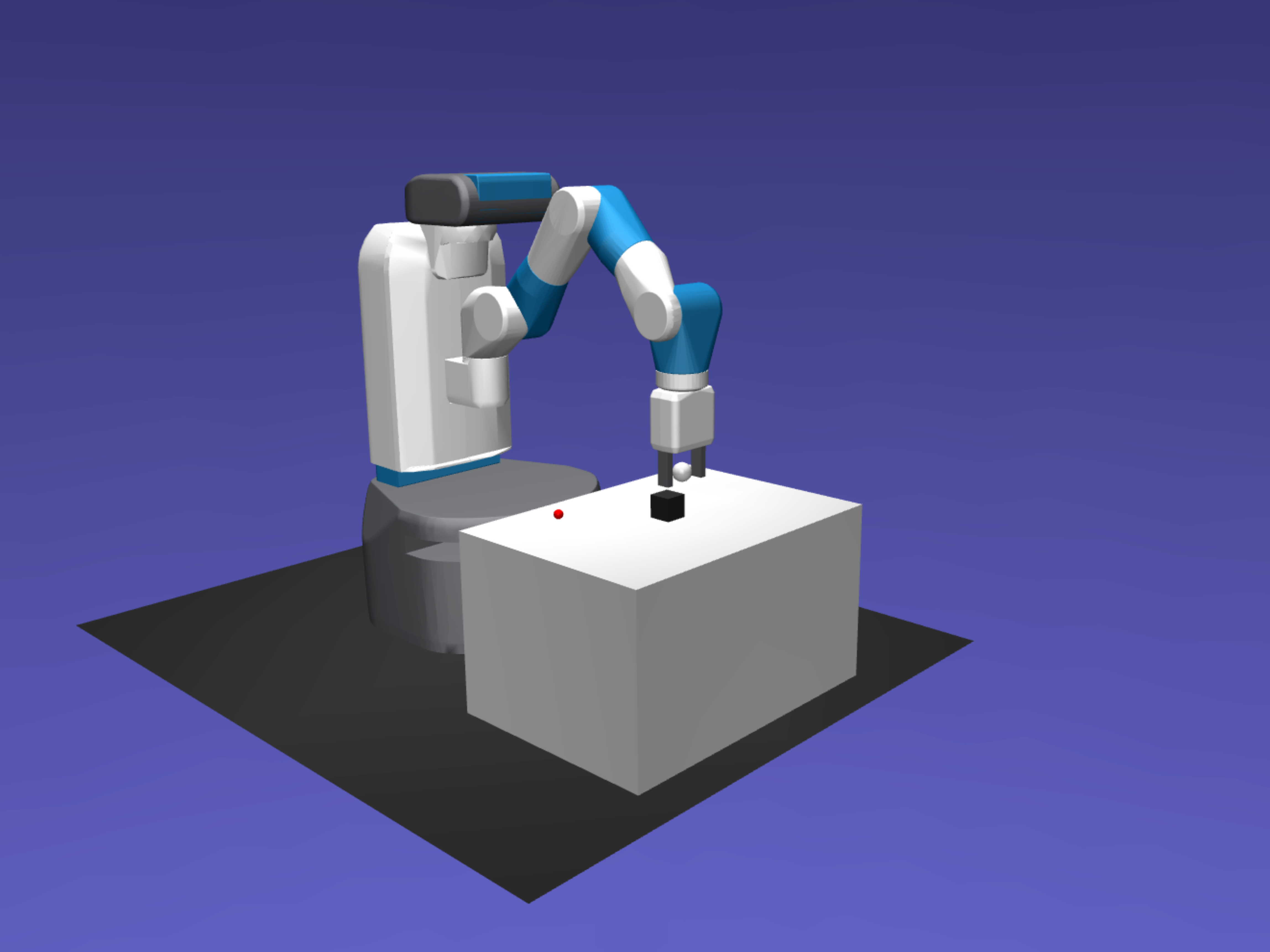} &
			\includegraphics[width=0.22\textwidth]{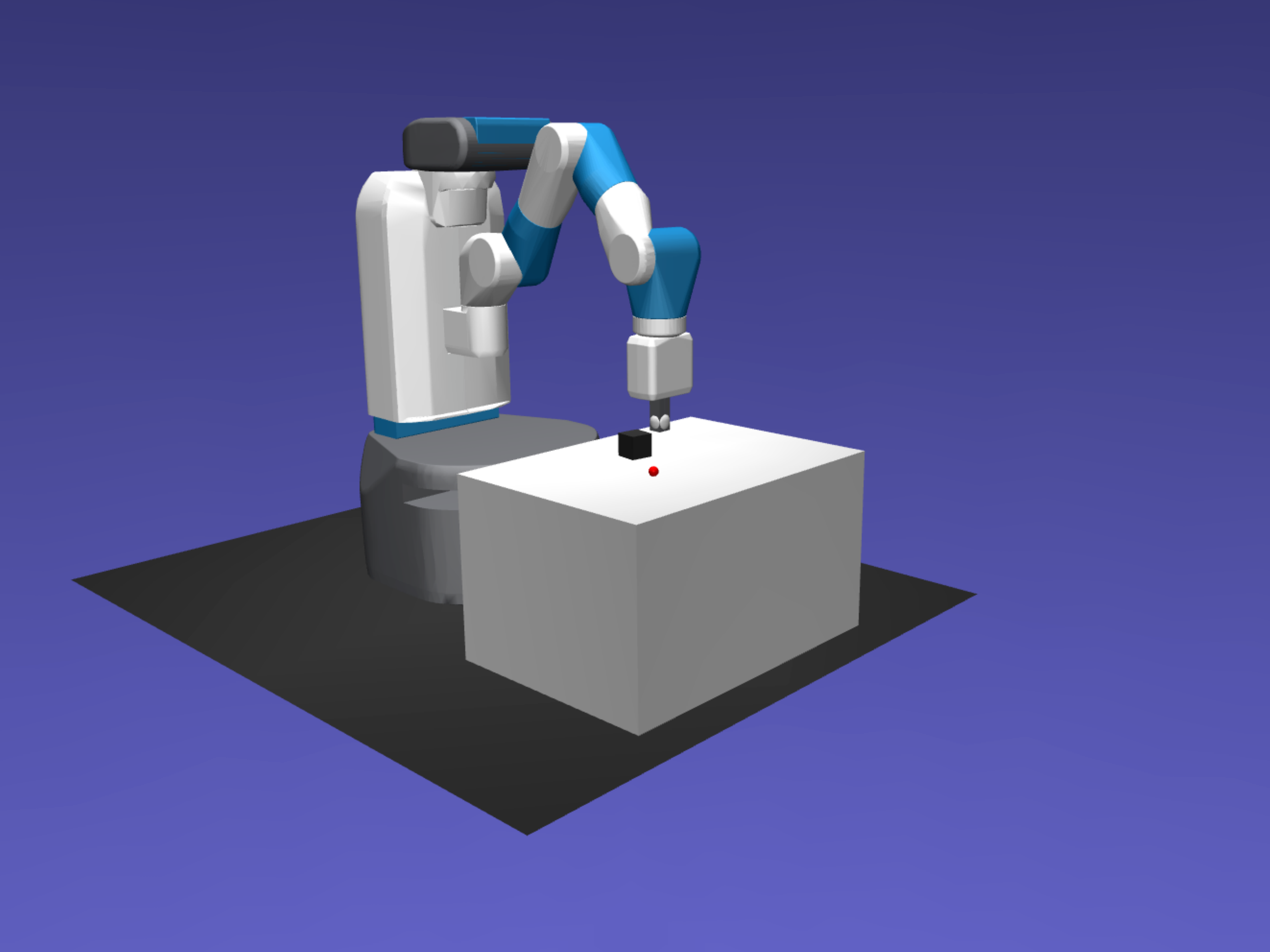} &   \includegraphics[width=0.22\textwidth]{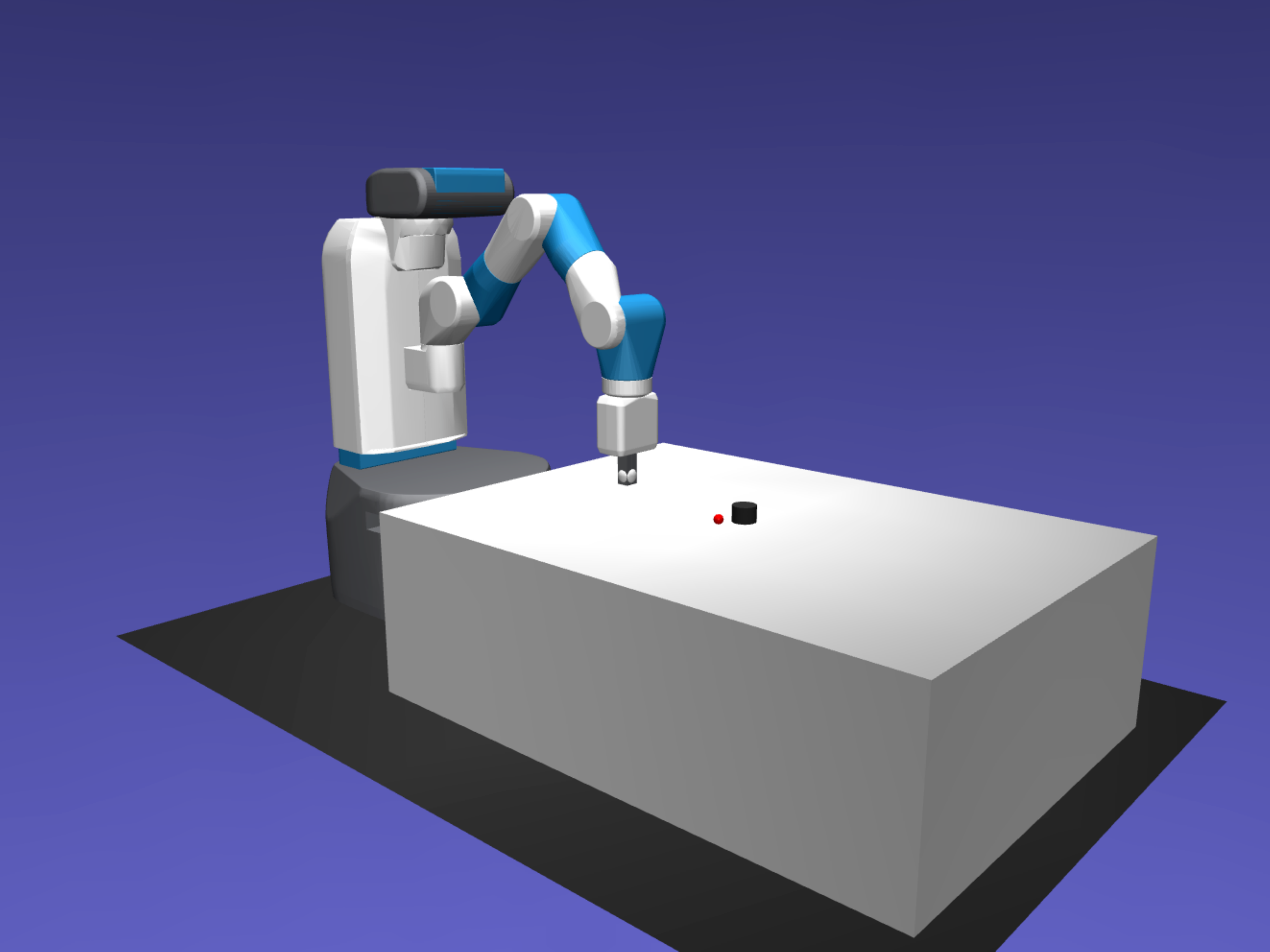} \\
			(a) & (b) & (c) & (d)
		\end{tabular}
		\caption{Robot arm Fetch environment for benchmark: (a) \texttt{FetchReach}, (b) \texttt{FetchPickAndPlace}, (c) \texttt{FetchPush}, and (d) \texttt{FetchSlide}.}
		\label{fig_chap_4:1}
	\end{figure}
	
	\begin{enumerate}
		\item \textbf{Reaching} (\texttt{FetchReach-v1}): The robot arm try to move its gripper from initial position to a desired target position. This target is located arbitrarily in 3D space.
		\item \textbf{Pick \& Place} (\texttt{FetchPickAndPlace-v1}): The task is to grasp a box and move it to the target location, which may be located on the surface of the table or in the air. In this task, the learned behavior is usually a mixture of grasping, slipping, and rolling.
		\item \textbf{Pushing} (\texttt{FetchPush-v1}): A box is placed randomly on a table in front of the robot, and the task is to move it to a target location on the table. The robot fingers are locked to prevent grasping. The learned behavior is usually a mixture of pushing and rolling.
		\item \textbf{Sliding} (\texttt{FetchSlide-v1}): A puck is placed on a long slippery table, and the target position is outside of the robot's reach so that it has to hit the puck with such a force that it slides and then stops at the target location due to the friction.
	\end{enumerate}
	
	The state is a vector consisting of the position, orientation, linear velocity, and angular velocity of all robot joints and objects. A goal represents the desired position and orientation of an object. There is an acceptable range around the desired position and direction. We use a fully sparse reward for all tasks like Eq. (\ref{sparsereward}) with a tolerance range of $\rho$ in our environment is 5 cm. If the object (or end-effector) is not in range of the goal, the agent receives a negative reward, ``-1"; otherwise, the positive reward ``0" is received.

	\noindent
	\textbf{Performance:} The proposed method is compared against the following baselines: Vanilla HER \cite{andrychowicz2017hindsight}, HER with Energy-Based Prioritization (EBP) \cite{zhao2018energy}, HER with Maximum Entropy-Regularized Prioritization (MEP) \cite{zhao2019maximum}, HER with one-step prioritization (PER) \cite{schaul2015prioritized,zhao2018energy}, and HER with HGR (Ours). We run the experiments in all four challenging object manipulation robotic environments with different random seeds. To evaluate sample efficiency, we compare how many samples the agent needs to achieve certain mean success rates at 50\%, 75\%, and 95\%. For \texttt{FetchSlide}, since it is hard to reach high performance without demonstration, we evaluate at 25\% and 45\% success rate. We also compare the final success rate at the end of training across methods. We evaluate ten times every epoch for each experiment, and the success rate is the average of ten times. Each experiment is averaged of five different random seeds, and the shaded area in the plot represents the standard deviation. In our experiment, we use 1 CPU instead of 19 CPUs as in previous work \cite{andrychowicz2017hindsight, zhao2018energy}.
	\footnote{If using 19 CPUs, the number of interactions with the environment increasing 19 times. In our work, the effectiveness of sampling for the single agent is verified.}.

	\subsection{Training details} \label{training_detail}
	We used identical architectures for actor and critic network- a three-layer network where each layer consists of 256 units and ReLU as the activation function. To train the HGR, the Adam \cite{kingma2014method} optimizer was used with a learning rate of $10^{-3}$ for both the actor and critic network. We updated the target Q function and target policy by using the moving average with the factor of 0.95. In the experiment, we used a discount factor $\gamma$ of 0.98. We used L2 regularization for action to prevent the predicted action from becoming too large. We used $ \tanh $ as the final layer's activation function such that the output value is scaled from -1 to 1. For exploration, we added a mean-zero Gaussian noise with 0.2 standard deviations to the deterministic policy. We also used the epsilon-greedy algorithm to alternate between uniform policy and behavior policy for better exploration with the factor $ \epsilon = 0.3 $. 
	
	\begin{table}[t]
		\caption{Hyperparameter used in our experiments. Most hyperparameters values are unchanged across environments with the exception for $ \alpha, \beta $, and training duration.} 
		\centering 
		\begin{tabular}{ccccc} 
			\hline\hline 
			Hyperparamerter & Reach         & PickPlace     & Push     & Slide \\ [0.5ex]
			\hline
			Buffer size   & 1e6  & 1e6       & 1e6     & 1e6 \\
			Batch size    & 256  & 256       & 256     & 256  \\
			Number of iterations     & 2e4   & 1.5e6      & 1e6     & 1e6 \\
			Actor learning rate      & 0.001 & 0.001 & 0.001 & 0.001 \\
			Critic learning rate     & 0.001 & 0.001 & 0.001 & 0.001 \\
			Discounted factor        & 0.98 & 0.98 & 0.98 & 0.98 \\
			Averaging coefficient    & 0.95 & 0.95 & 0.95 & 0.95 \\
			$\alpha=\alpha'$         & 0.6 & 0.6 & 0.8 & 0.6 \\
			$\beta=\beta'$           & 0.4 & 0.4 & 0.5 & 0.4 \\
			Gaussian noise std  ($\sigma$)     & 0.2 & 0.2 & 0.2 & 0.2 \\
			$\epsilon$-greedy factor    & 0.3 & 0.3 & 0.3 & 0.3 \\
			Action L2 regularization  ($\sigma$)     & 1.0 & 1.0 & 1.0 & 1.0 \\
			\hline 
		\end{tabular}
		\label{summarized_hyperparameters}
	\end{table}
	
	To determine the hyperparameters $ \alpha, \alpha', \beta, $ and $ \beta' $ for prioritization,  we performed coarse grid search with range of values as $\alpha = \{0, 0.4, 0.5, 0.6, 0.7, 0.8\}, \beta =\{0, 0.4, 0.5, 0.6, 0.7, 1\}$.  We found $ \alpha=\alpha' = 0.6, \beta=\beta' = 0.4 $ led to best performance for \texttt{FetchReach}, \texttt{FetchPickAndPlace}, and \texttt{FetchSlide} task. For \texttt{FetchPush},  $ \alpha=\alpha' = 0.8,  \beta=\beta' = 0.5 $ gave best performance. We kept all hyper-parameter fixed for all the experiments. The buffer size for all environments was set to 1e6. For \texttt{FetchReach}, the agent interacted with the environment 2e4 times, and for\texttt{FetchPush}, \texttt{FetchSlide}, the agent interacted for 1e6 times, and for \texttt{FetchPickAndPlace}, it interacted 1.5e6 times. For example with horizon $H=50$, there are 1.5e6/50=30,000 episodes for \texttt{FetchPickAndPlace}. The summary of the hyperparameters is shown in Table \ref{summarized_hyperparameters}.

	\subsection{Compare with recent methods}
	
\begin{figure}[t]
		\centering
		\begin{tabular}{cc}
			\includegraphics[width=0.44\textwidth]{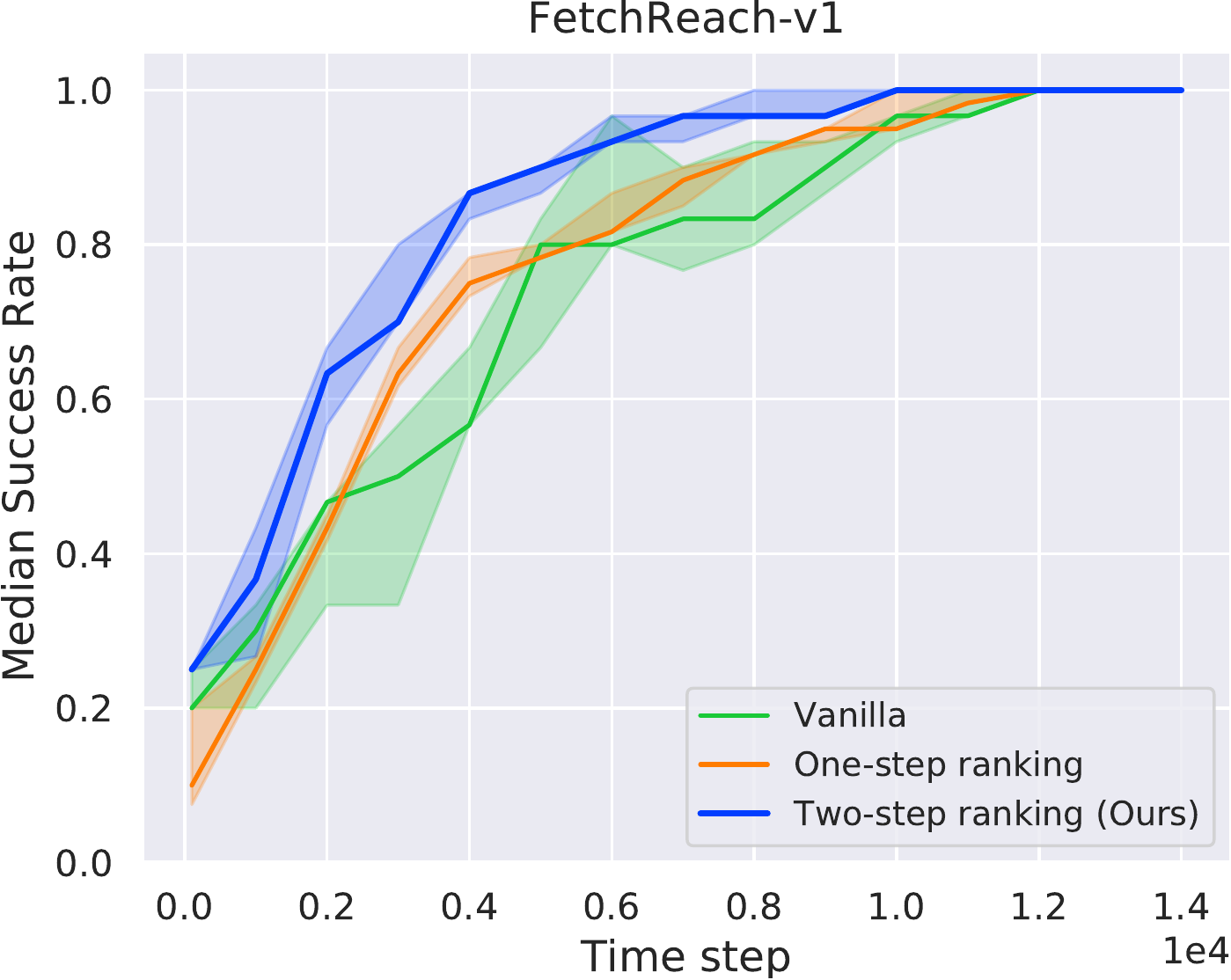} &   \includegraphics[width=0.44\textwidth]{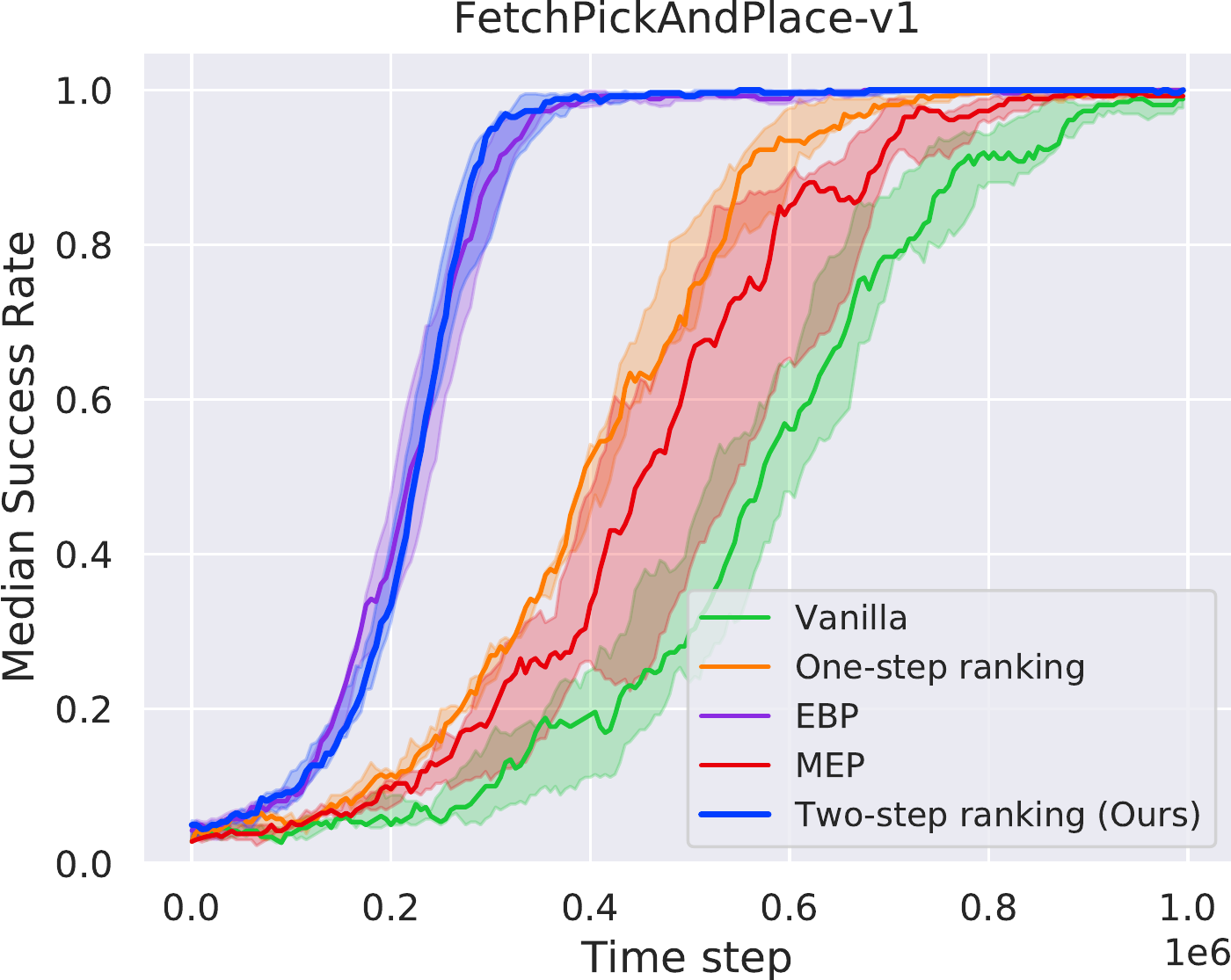} \\
			\includegraphics[width=0.44\textwidth]{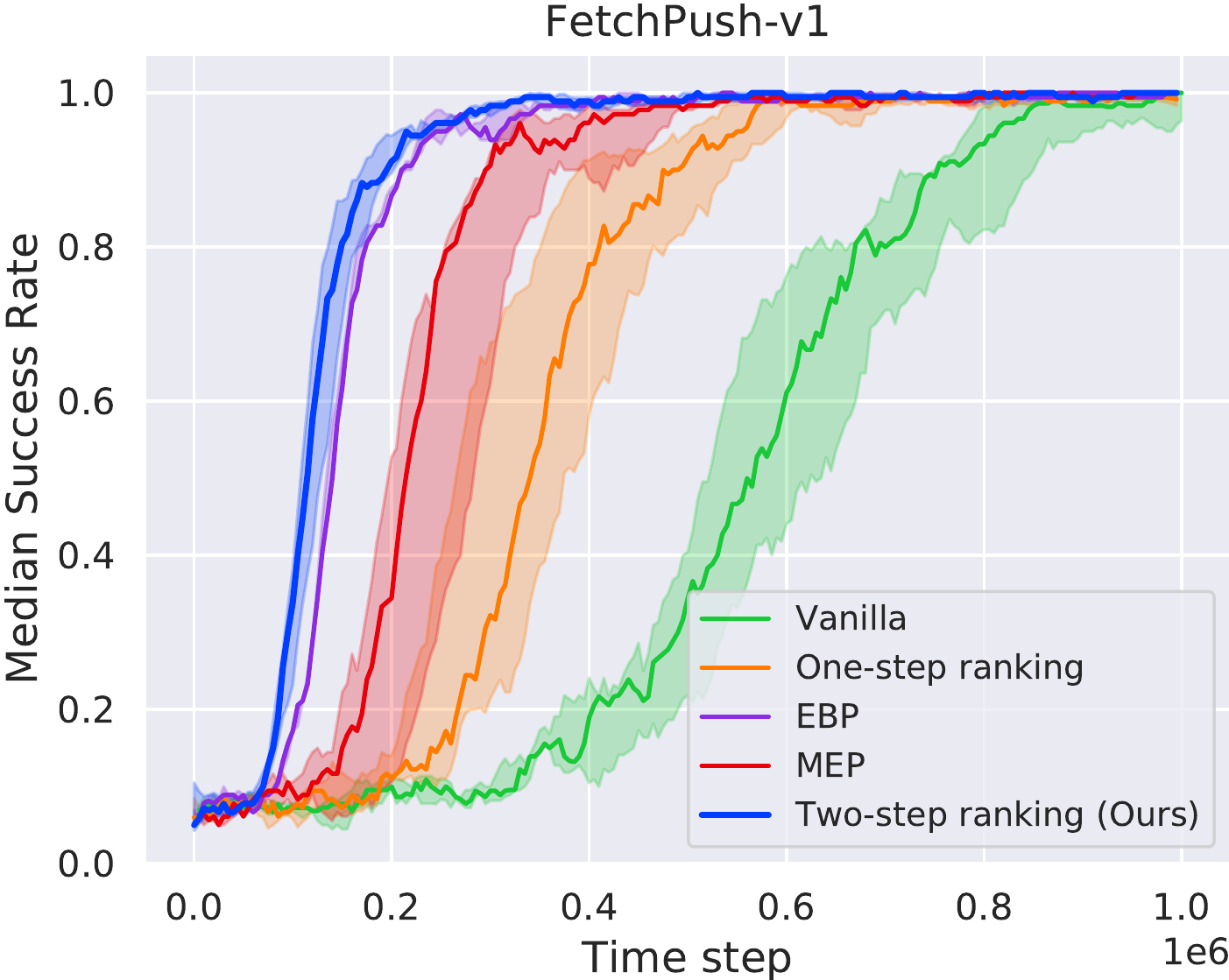} &   \includegraphics[width=0.44\textwidth]{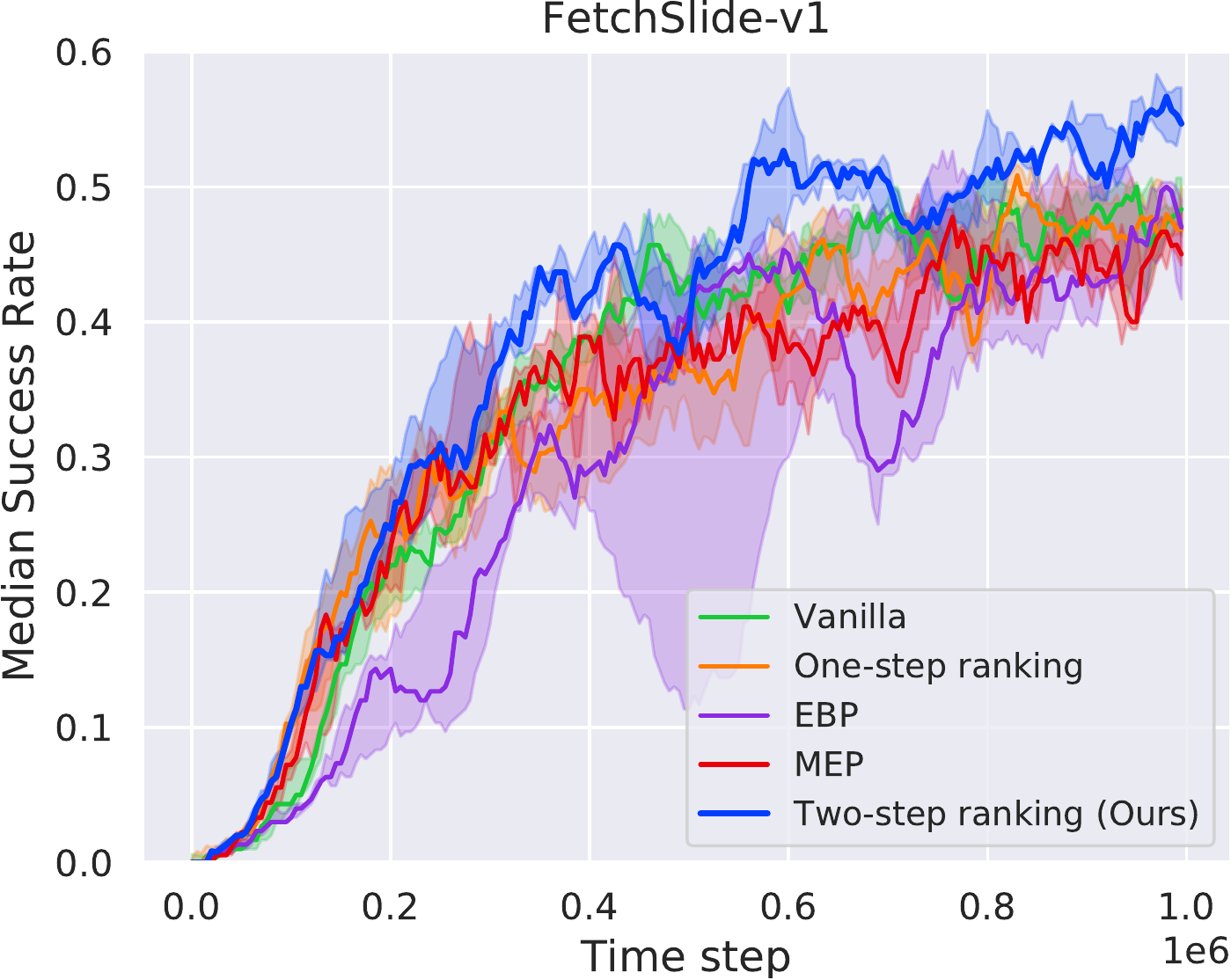}
		\end{tabular}
		\caption{The evolution of mean test success rate across five seeds with standard deviation for all four tasks simulated on Fetch robotic. Overall, DDPG combined with HGR shows the use of replay buffer more sample efficient, with $ 2.9 \times $ faster than Vanilla HER, $ 1.3 \times $ and $ 1.8 \times $ faster than EBP \cite{zhao2018energy} and MEP \cite{zhao2019maximum}, respectively.}
		\label{fig_chap_4:2}
	\end{figure}

	In the considered tasks, the goal can be changed to an arbitrary location when beginning a new episode and kept fixed within the episode. Hence, to solve these tasks, the RL algorithms must have the ability to adapt to multiple goals. Figure \ref{fig_chap_4:2} shows evolution of success rate in four benchmark environments. For comparison with baseline vanilla HER and HER with PER, the result shows that HER combined with HGR converges significantly faster in all four tasks. For comparison with Energy-Based Prioritization (EBP) \footnote{We used author's code at this link:  \url{https://github.com/ruizhaogit/EnergyBasedPrioritization}}, we achieve better in \texttt{FetchSlide, FetchPush}, and slightly better in \texttt{FetchPickAndPlace}. For HER with Maximum-Regularized Entropy Prioritization (MEP), our method surpasses across all environments.
	
	\begin{table}[t]
		\caption{The number of samples required to achieve a certain success rate (lower it better, `$-$' means not support for this environment). The results are recorded every epoch and averaged across five random seeds. }
		\centering
		\begin{tabular}{cccc|ccc|ccc|cc} 
			\hline\hline 
			& \multicolumn{3}{c}{Reach ($ \times 1e4 $)} & \multicolumn{3}{c}{PickPlace ($ \times 1e6 $)} & \multicolumn{3}{c}{Push ($ \times 1e6 $)} & \multicolumn{2}{c}{Slide ($ \times 1e6 $)} \\
			\hline
			Success & 50\% 	  & 75\% 	& 95\% 	  & 50\% 	  & 75\% 	  & 95\% 	 & 50\% 	& 75\% 		& 95\% 	   & 25\% 	  & 45\% \\
			\hline 
			Vanilla &0.4 	  &0.5  	&1.0 	  & 0.585 	  & 0.750 	  &1.180   	 & 0.565 	& 0.755 	& 0.925	   & 0.260 	  &0.665 \\ 
			PER     &0.3	  &0.5		&1.0 	  & 0.375	  & 0.480	  &0.625 	 & 0.335 	& 0.410		& 0.515    & 0.215 	  &0.710 \\ 
			EBP     &$ - $&$ - $ 	&$ - $ 	  & \bf{0.220}& 0.270  	  &0.340 	 & 0.145  	& 0.175  	& 0.225    & 0.330 	  &0.930 \\
			MEP     &$ - $ 	  &$ - $ 	&$ - $ 	  & 0.470  	  & 0.585  	  &0.790 	 & 0.235  	& 0.300  	& 0.370    & 0.215 	  &0.755 \\
			Ours    &\bf{0.2}&\bf{0.3}&\bf{0.7}& \bf{0.220}& \bf{0.260}&\bf{0.335}&\bf{0.115}& \bf{0.145}&\bf{0.190}&\bf{0.200}&\bf{0.545} \\ 
			\hline 
		\end{tabular}
		\label{evolution_performance}
	\end{table}

	Specifically, Table \ref{evolution_performance} shows the number of samples to achieve a certain performance. In \texttt{FetchReach}, ours method requires 2000, 3000, and 7000 samples to reach 50\%, 75\%, and 95\% success rate, respectively. In both three levels, ours converge faster than HER+PER and Vanilla HER. In \texttt{FetchPickAndPlace}, HER+HGR achieves 50\%, 75\%, and 95\% faster success rate with 220000, 260000, and 335000 samples, respectively. Overall, HER+HGR converges roughly $ 3\times $ faster than Vanilla HER, $ 1.8\times, 2.2\times $ faster than HER+PER and HER+MEP, and slightly faster than HER+EBP. In \texttt{FetchPush}, HER+HGR needs 115000, 145000, and 190000 samples, which is $ 5\times, 2.8\times, 2.0\times $, and $ 1.2\times $ less than Vanilla HER, PER, MEP, and EBP, respectively. In \texttt{FetchSlide}, without demonstrations, ours needs 200000 and 545000 samples to reach to 25\% and 45\% success rate, which is more efficient than the others.

	The final performance of the trained agent is shown in Table \ref{finalperformance}. In \texttt{Reach}, both three methods can achieve absolutely performance. For the other three environments, HER+HGR outperforms other methods. The improvement varies from 0.5 percent point to 15.8 percent point compared to Vanilla HER, from 0.4 percent to 8.9 percent point compared to PER. For comparison with state of the art EBP, our method is outperformed in \texttt{FetchPickAndPlace}, \texttt{FetchPush}, and \texttt{FetchSlide} with 0.07, 0.15, and 8.4 percent point, respectively. Overall, our two-step ranking method benefits HER and shows the use of samples more effective than baseline methods in the robotic environments. However, ranking the relay buffer requires overhead. Overall, the training time is roughly three times slower compared to Energy-Based Prioritization and four times compared to Vanilla HER.
	
	\subsection{Ablation studies}
	
	\begin{figure}[t]
		\centering
		\begin{tabular}{cc}
			\includegraphics[width=0.44\textwidth]{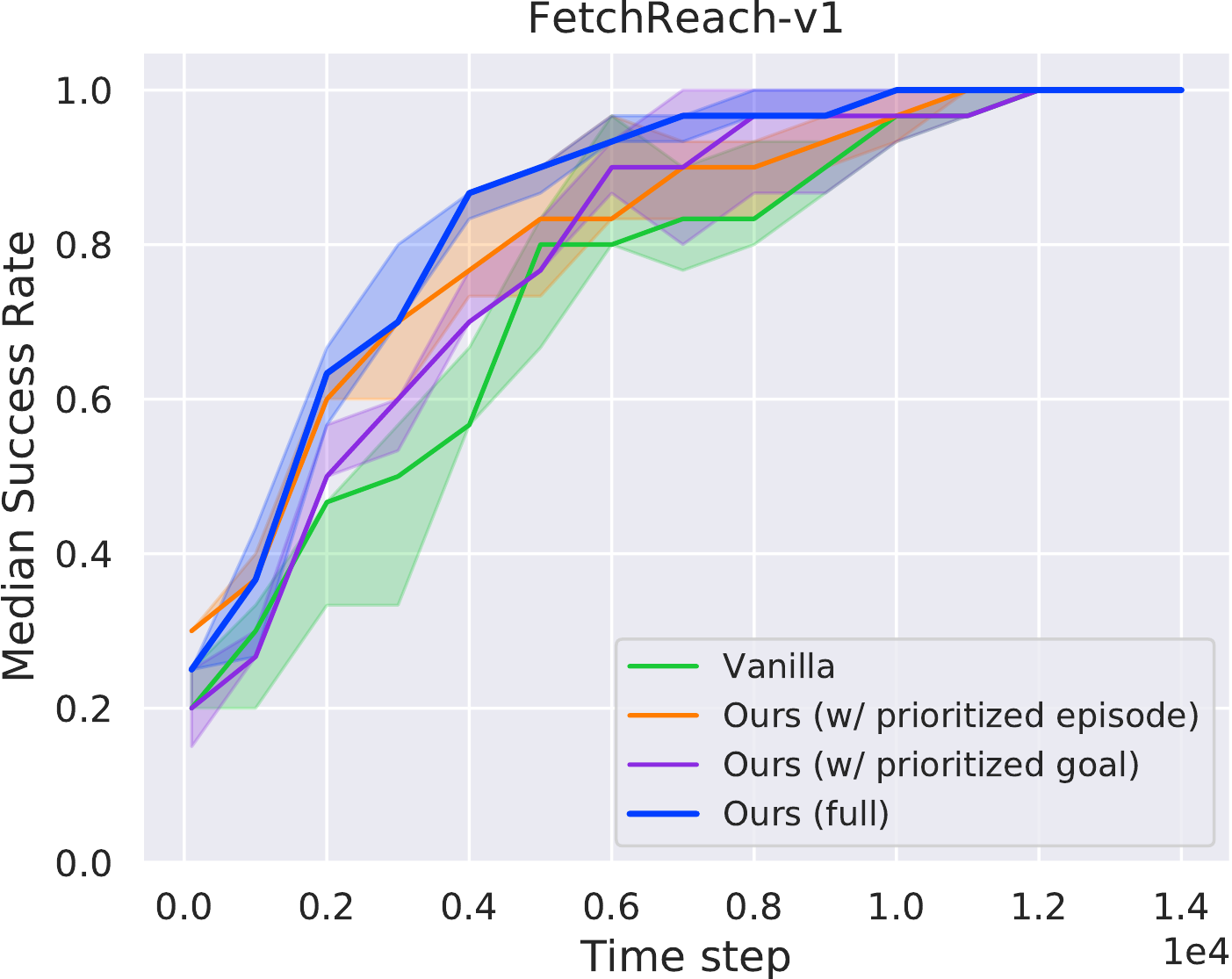} &   \includegraphics[width=0.44\textwidth]{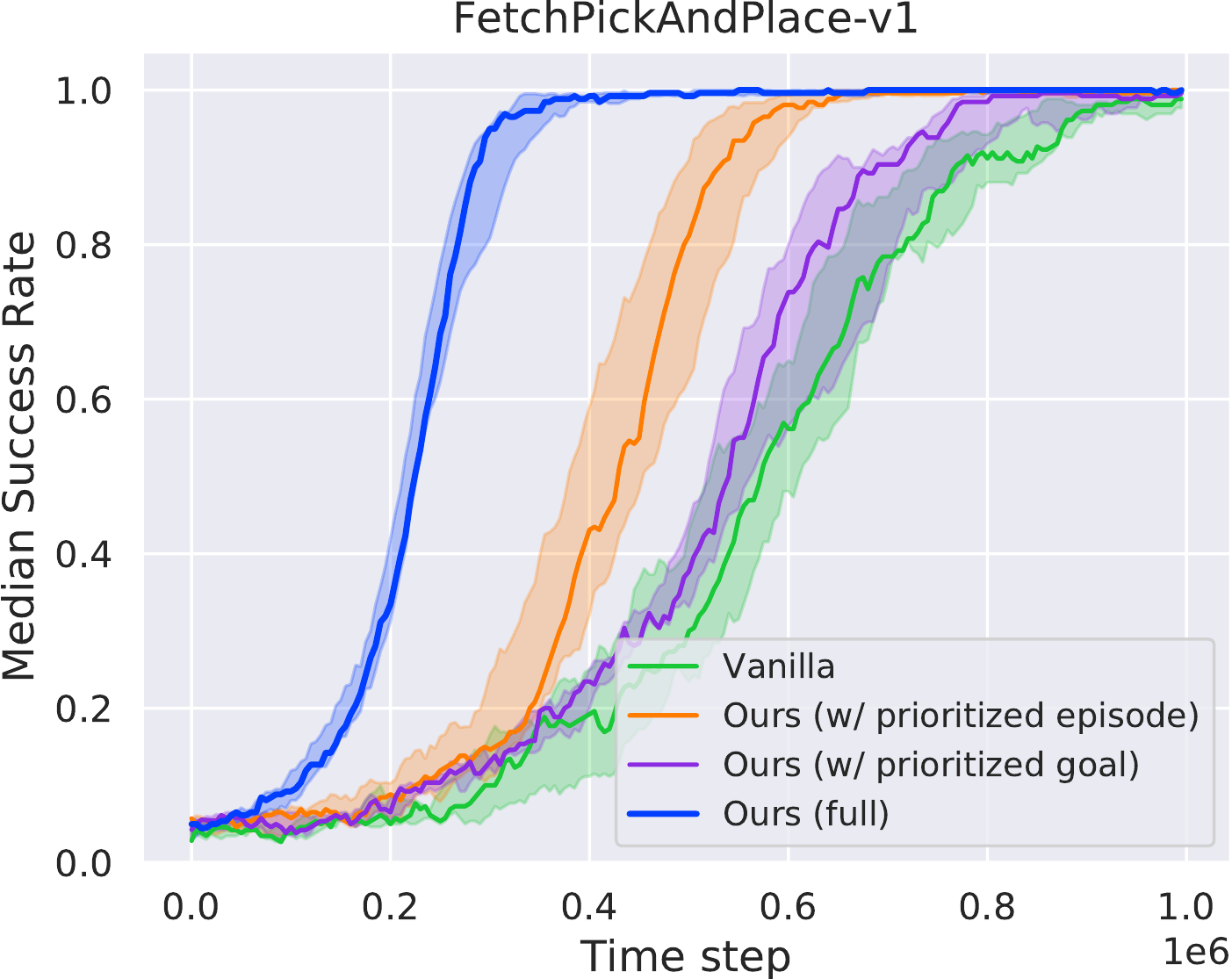} \\
			\includegraphics[width=0.44\textwidth]{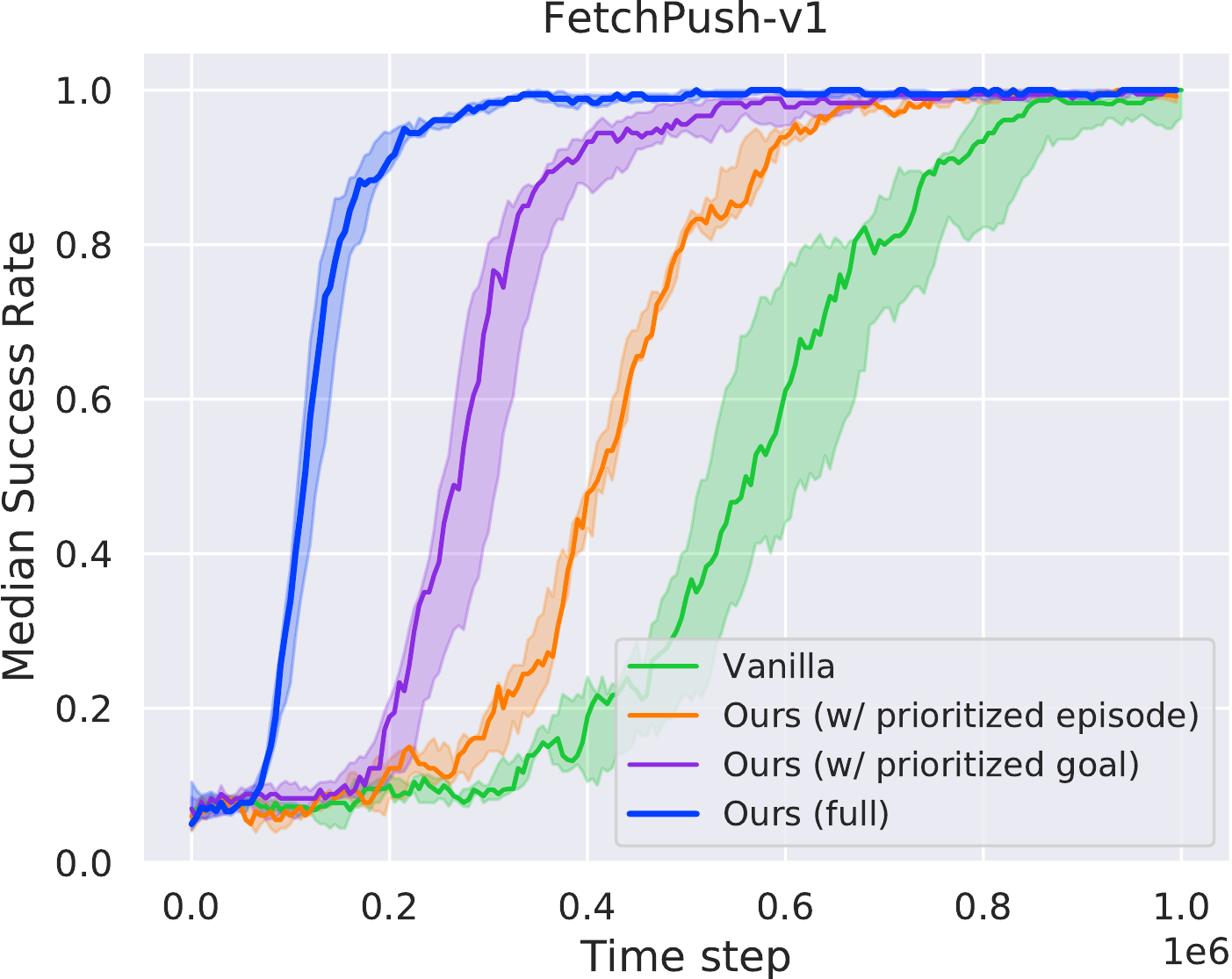} & 
			\includegraphics[width=0.44\textwidth]{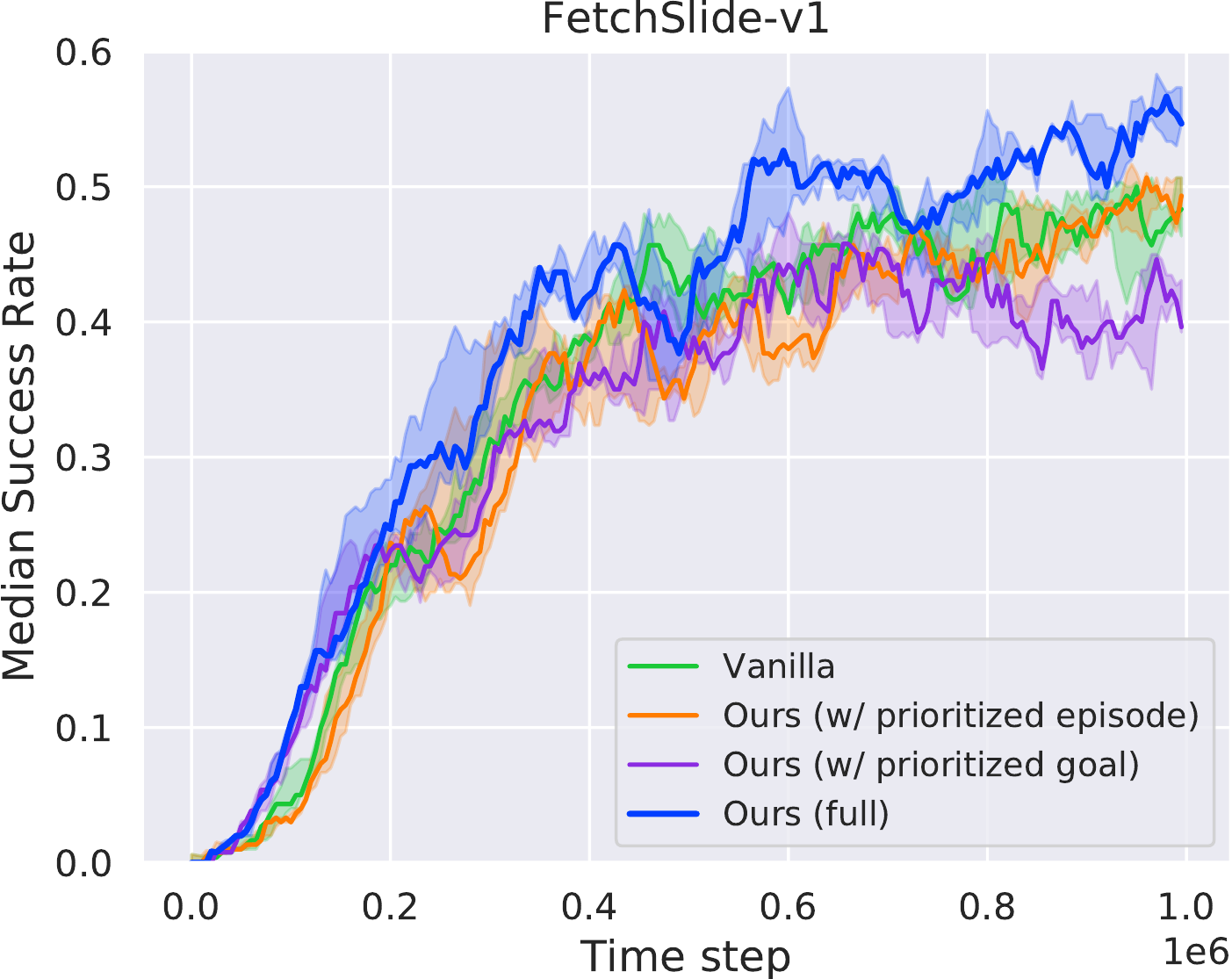}
		\end{tabular}
		\caption{Ablation study for three environments: ``full'' means using two-step ranking, ``w/ prioritized goal'' means only ranking goals, uniformly sampling episodes, ``w/ prioritized goal'' means only ranking episode, uniformly sampling goals.}
		\label{fig_chap_4:3}
	\end{figure}
	
	\begin{table}[t]
		\caption{The mean success rate on four manipulation tasks. The results are recorded at the end of training and averaged across five random seeds. On each task we mark as bold the highest score. Our method demonstrates better overall performance.}
		\centering 
		\begin{tabular}{c c c c c} 
			\hline\hline 
			& Reach         & PickPlace     & Push     	& Slide \\
			\hline
			Vanilla   & \bf 100\%  & 99.50\%       & 93.00\%     & 41.10\% \\
			PER       & \bf 100\%  & 99.60\%       & 99.36\%     & 48.00\% \\
			EBP       & $-$  		  & 99.93\%       & 99.71\%     & 48.50\% \\
			MEP       & $-$  	 	  & 99.80\%       & 99.80\%     & 45.40\% \\
			Ours      & \bf 100\%  & \bf{100}\%  &\bf{99.86}\% & \bf{56.90\%} \\
			\hline 
		\end{tabular}
		\label{finalperformance}
	\end{table}

	To investigate the benefit between goal prioritization and episode prioritization, ablation studies on all four benchmark environments \texttt{FetchReach}, \texttt{FetchPush}, \texttt{FetchPickAndPlace}, and \texttt{FetchSlide} are conducted. To disable episode prioritization, we set $ \alpha=0 $ and $ \beta=0 $, and to disable goal prioritization, we set $ \alpha'=0 $ and $ \beta'=0 $. The evolution of success rate using different prioritization types is shown in Figure \ref{fig_chap_4:3}. In the \texttt{FetchReach} and \texttt{FetchSlide}, the benefit of each ranking is unclear as 
	the former is an easy task to learn, and the latter is more difficult without supporting of demonstrations.
	In \texttt{FetchReach}, each type of prioritization led to better performance than vanilla HER. In \texttt{FetchSlide}, the ranking with the prioritized goal is even worst than no prioritization, but the combination of the two steps achieves better results. In \texttt{FetchPush}, at the same $ 100^{th} $ episode corresponding one million time step, the success rate of ``full", ``prioritized goal", ``prioritized episode", and without any prioritization are 99.05, 89.74, 71.70 and 31.59 percent, respectively. This indicates that ranking the goal is more important than ranking the episode in the \texttt{FetchPush} environment. In contrast, the episode's prioritization in \texttt{FetchPickAndPlace} is more important. Specifically, at $ 100^{th} $ episode, in the \texttt{FetchPickAndPlace} environment, the success rate of ``full", ``prioritized goal", ``prioritized episode", and without any prioritization are 99.40, 43.25, 70.13 and 27.02 percent, respectively. Overall, in comparison with Vanilla HER, the two-step ranking HGR at the episode as well as at the goal level improves sample efficiency and final performance. How much of the benefit would depend on the task.
	
	\section{Conclusion} \label{section:6}
	
	In this paper, a prioritized replay method for multi-goal setting in the sparse rewards environment is considered. Inspired by the prioritization method in PER, which is proposed for a single goal and discrete action space, prioritization for multi-goal and continuous action space environment is studied. The proposed method divides the prioritized sampling into two steps: first, an episode is sampled according to the average TD error of experience with hindsight goals within the episode, then, for the sampled episode, experience with hindsight goals leading to larger TD error is sampled with higher probability. From the empirical results, HER with HGR significantly improves sample efficiency in the multi-goal RL with the sparse reward environment compared with vanilla HER, and its performance is marginally higher performance than the state of the art algorithms. However, HER with HGR requires $ O(\log n) $ computation times to search and update the priority.

	\section*{Acknowledgment}
	This work was supported in part by the Institute for Information \& Communications Technology Planning \& Evaluation (IITP) grant funded by the Korea Government (MSIT) (No. 2019-0-01396, Development of Framework for Analyzing, Detecting, Mitigating of Bias in AI Model and Training Data) and in part by the BK21 FOUR program.

	
	\bibliographystyle{plain}
	\bibliography{main}

\begin{thebibliography}{10}

\bibitem{andre1998generalized}
David Andre, Nir Friedman, and Ronald Parr.
\newblock Generalized prioritized sweeping.
\newblock In {\em Advances in Neural Information Processing Systems (NeurIPS)},
  pages 1001--1007, 1998.

\bibitem{andrychowicz2017hindsight}
Marcin Andrychowicz, Filip Wolski, Alex Ray, Jonas Schneider, Rachel Fong,
  Peter Welinder, Bob McGrew, Josh Tobin, OpenAI~Pieter Abbeel, and Wojciech
  Zaremba.
\newblock Hindsight experience replay.
\newblock In {\em Advances in Neural Information Processing Systems (NeurIPS)},
  pages 5048--5058, 2017.

\bibitem{1606.01540}
Greg Brockman, Vicki Cheung, Ludwig Pettersson, Jonas Schneider, John Schulman,
  Jie Tang, and Wojciech Zaremba.
\newblock Openai gym, 2016.

\bibitem{colas2019curious}
C{\'e}dric Colas, Pierre Fournier, Mohamed Chetouani, Olivier Sigaud, and
  Pierre-Yves Oudeyer.
\newblock Curious: Intrinsically motivated modular multi-goal reinforcement
  learning.
\newblock In {\em International conference on machine learning (ICML)}, pages
  1331--1340, 2019.

\bibitem{ding2019goal}
Yiming Ding, Carlos Florensa, Pieter Abbeel, and Mariano Phielipp.
\newblock Goal-conditioned imitation learning.
\newblock In {\em Advances in Neural Information Processing Systems (NeurIPS)},
  pages 15324--15335, 2019.

\bibitem{duan2016benchmarking}
Yan Duan, Xi~Chen, Rein Houthooft, John Schulman, and Pieter Abbeel.
\newblock Benchmarking deep reinforcement learning for continuous control.
\newblock In {\em International Conference on Machine Learning (ICML)}, pages
  1329--1338, 2016.

\bibitem{eysenbach2020rewriting}
Benjamin Eysenbach, Xinyang Geng, Sergey Levine, and Ruslan Salakhutdinov.
\newblock Rewriting history with inverse rl: Hindsight inference for policy
  improvement.
\newblock {\em arXiv preprint arXiv:2002.11089}, 2020.

\bibitem{fang2018dher}
Meng Fang, Cheng Zhou, Bei Shi, Boqing Gong, Jia Xu, and Tong Zhang.
\newblock Dher: Hindsight experience replay for dynamic goals.
\newblock In {\em International Conference on Learning Representations (ICML)},
  2019.

\bibitem{florensa2018automatic}
Carlos Florensa, David Held, Xinyang Geng, and Pieter Abbeel.
\newblock Automatic goal generation for reinforcement learning agents.
\newblock In Jennifer Dy and Andreas Krause, editors, {\em Proceedings of the
  35th International Conference on Machine Learning}, volume~80 of {\em
  Proceedings of Machine Learning Research}, pages 1515--1528,
  Stockholmsmässan, Stockholm Sweden, 10--15 Jul 2018. PMLR.

\bibitem{florensa2017reverse}
Carlos Florensa, David Held, Markus Wulfmeier, Michael Zhang, and Pieter
  Abbeel.
\newblock Reverse curriculum generation for reinforcement learning.
\newblock In Sergey Levine, Vincent Vanhoucke, and Ken Goldberg, editors, {\em
  Proceedings of the 1st Annual Conference on Robot Learning}, volume~78 of
  {\em Proceedings of Machine Learning Research}, pages 482--495. PMLR, 13--15
  Nov 2017.

\bibitem{forestier2017intrinsically}
S{\'e}bastien Forestier, Yoan Mollard, and Pierre-Yves Oudeyer.
\newblock Intrinsically motivated goal exploration processes with automatic
  curriculum learning.
\newblock {\em arXiv preprint arXiv:1708.02190}, 2017.

\bibitem{fujimoto2018addressing}
Scott Fujimoto, Herke Van~Hoof, and David Meger.
\newblock Addressing function approximation error in actor-critic methods.
\newblock {\em International Conference on Machine Learning (ICML)}, 2018.

\bibitem{haarnoja2018soft}
Tuomas Haarnoja, Aurick Zhou, Pieter Abbeel, and Sergey Levine.
\newblock Soft actor-critic: Off-policy maximum entropy deep reinforcement
  learning with a stochastic actor.
\newblock {\em International Conference on Machine Learning (ICML)}, 2018.

\bibitem{ho2016generative}
Jonathan Ho and Stefano Ermon.
\newblock Generative adversarial imitation learning.
\newblock In {\em Advances in neural information processing systems (NeurIPS)},
  pages 4565--4573, 2016.

\bibitem{kalashnikov2018qt}
Dmitry Kalashnikov, Alex Irpan, Peter Pastor, Julian Ibarz, Alexander Herzog,
  Eric Jang, Deirdre Quillen, Ethan Holly, Mrinal Kalakrishnan, Vincent
  Vanhoucke, et~al.
\newblock Qt-opt: Scalable deep reinforcement learning for vision-based robotic
  manipulation.
\newblock {\em Conference on Robot Learning (CoRL)}, 2018.

\bibitem{kim2004autonomous}
H~Jin Kim, Michael~I Jordan, Shankar Sastry, and Andrew~Y Ng.
\newblock Autonomous helicopter flight via reinforcement learning.
\newblock In {\em Advances in neural information processing systems (NeurIPS)},
  pages 799--806, 2004.

\bibitem{kingma2014method}
Diederik~P. Kingma and Jimmy Ba.
\newblock Adam: A method for stochastic optimization, 2014.
\newblock Published as a conference paper at the 3rd International Conference
  for Learning Representations, San Diego, 2015.

\bibitem{levine2016end}
Sergey Levine, Chelsea Finn, Trevor Darrell, and Pieter Abbeel.
\newblock End-to-end training of deep visuomotor policies.
\newblock {\em The Journal of Machine Learning Research}, 17(1):1334--1373,
  2016.

\bibitem{li2020generalized}
Alexander~C Li, Lerrel Pinto, and Pieter Abbeel.
\newblock Generalized hindsight for reinforcement learning.
\newblock {\em arXiv preprint arXiv:2002.11708}, 2020.

\bibitem{lillicrap2015continuous}
Timothy~P Lillicrap, Jonathan~J Hunt, Alexander Pritzel, Nicolas Heess, Tom
  Erez, Yuval Tassa, David Silver, and Daan Wierstra.
\newblock Continuous control with deep reinforcement learning.
\newblock {\em International conference on machine learning (ICML)}, 2016.

\bibitem{lin1992self}
Long-Ji Lin.
\newblock Self-improving reactive agents based on reinforcement learning,
  planning and teaching.
\newblock {\em Machine learning}, 8(3-4):293--321, 1992.

\bibitem{mnih2013playing}
Volodymyr Mnih, Koray Kavukcuoglu, David Silver, Alex Graves, Ioannis
  Antonoglou, Daan Wierstra, and Martin Riedmiller.
\newblock Playing atari with deep reinforcement learning.
\newblock {\em Advances in Neural Information Processing Systems (NeurIPS) Deep
  Learning Workshop}, 2013.

\bibitem{mnih2015human}
Volodymyr Mnih, Koray Kavukcuoglu, David Silver, Andrei~A Rusu, Joel Veness,
  Marc~G Bellemare, Alex Graves, Martin Riedmiller, Andreas~K Fidjeland, Georg
  Ostrovski, et~al.
\newblock Human-level control through deep reinforcement learning.
\newblock {\em Nature}, 518(7540):529, 2015.

\bibitem{nair2018overcoming}
Ashvin Nair, Bob McGrew, Marcin Andrychowicz, Wojciech Zaremba, and Pieter
  Abbeel.
\newblock Overcoming exploration in reinforcement learning with demonstrations.
\newblock In {\em 2018 IEEE International Conference on Robotics and Automation
  (ICRA)}, pages 6292--6299. IEEE, 2018.

\bibitem{nair2018visual}
Ashvin~V Nair, Vitchyr Pong, Murtaza Dalal, Shikhar Bahl, Steven Lin, and
  Sergey Levine.
\newblock Visual reinforcement learning with imagined goals.
\newblock In {\em Advances in Neural Information Processing Systems (NeurIPS)},
  pages 9191--9200, 2018.

\bibitem{ng2006autonomous}
Andrew~Y Ng, Adam Coates, Mark Diel, Varun Ganapathi, Jamie Schulte, Ben Tse,
  Eric Berger, and Eric Liang.
\newblock Autonomous inverted helicopter flight via reinforcement learning.
\newblock In {\em Experimental robotics IX}, pages 363--372. Springer, Berlin,
  2006.

\bibitem{plappert2018multi}
Matthias Plappert, Marcin Andrychowicz, Alex Ray, Bob McGrew, Bowen Baker,
  Glenn Powell, Jonas Schneider, Josh Tobin, Maciek Chociej, Peter Welinder,
  et~al.
\newblock Multi-goal reinforcement learning: Challenging robotics environments
  and request for research.
\newblock {\em arXiv preprint arXiv:1802.09464}, 2018.

\bibitem{pomerleau1989alvinn}
Dean~A Pomerleau.
\newblock Alvinn: An autonomous land vehicle in a neural network.
\newblock In {\em Advances in neural information processing systems (NeurIPS)},
  pages 305--313, 1989.

\bibitem{pong2019skew}
Vitchyr~H Pong, Murtaza Dalal, Steven Lin, Ashvin Nair, Shikhar Bahl, and
  Sergey Levine.
\newblock Skew-fit: State-covering self-supervised reinforcement learning.
\newblock {\em arXiv preprint arXiv:1903.03698}, 2019.

\bibitem{rauber2019hindsight}
Paulo Rauber, Avinash Ummadisingu, Filipe Mutz, and Juergen Schmidhuber.
\newblock Hindsight policy gradients.
\newblock {\em International conference on machine learning (ICML)}, 2019.

\bibitem{schaul2015universal}
Tom Schaul, Daniel Horgan, Karol Gregor, and David Silver.
\newblock Universal value function approximators.
\newblock In {\em International conference on machine learning (ICML)}, pages
  1312--1320, 2015.

\bibitem{schaul2015prioritized}
Tom Schaul, John Quan, Ioannis Antonoglou, and David Silver.
\newblock Prioritized experience replay.
\newblock {\em International Conference on Learning Representations (ICLR)},
  2016.

\bibitem{silver2016mastering}
David Silver, Aja Huang, Chris~J Maddison, Arthur Guez, Laurent Sifre, George
  Van Den~Driessche, Julian Schrittwieser, Ioannis Antonoglou, Veda
  Panneershelvam, Marc Lanctot, et~al.
\newblock Mastering the game of go with deep neural networks and tree search.
\newblock {\em Nature}, 529(7587):484, 2016.

\bibitem{silver2017mastering}
David Silver, Thomas Hubert, Julian Schrittwieser, Ioannis Antonoglou, Matthew
  Lai, Arthur Guez, Marc Lanctot, Laurent Sifre, Dharshan Kumaran, Thore
  Graepel, et~al.
\newblock Mastering chess and shogi by self-play with a general reinforcement
  learning algorithm.
\newblock {\em arXiv preprint arXiv:1712.01815}, 2017.

\bibitem{silver2014deterministic}
David Silver, Guy Lever, Nicolas Heess, Thomas Degris, Daan Wierstra, and
  Martin Riedmiller.
\newblock Deterministic policy gradient algorithms.
\newblock In {\em International Conference on Machine Learning (ICML)}, 2014.

\bibitem{sutton2018reinforcement}
Richard~S Sutton and Andrew~G Barto.
\newblock {\em Reinforcement learning: An introduction}.
\newblock MIT press, 2018.

\bibitem{mujoco}
E.~{Todorov}, T.~{Erez}, and Y.~{Tassa}.
\newblock Mujoco: A physics engine for model-based control.
\newblock In {\em 2012 IEEE/RSJ International Conference on Intelligent Robots
  and Systems}, pages 5026--5033, Oct 2012.

\bibitem{van2016deep}
Hado Van~Hasselt, Arthur Guez, and David Silver.
\newblock Deep reinforcement learning with double q-learning.
\newblock In {\em Association for the Advancement of Artificial Intelligence
  (AAAI)}, 2016.

\bibitem{wang2015dueling}
Ziyu Wang, Tom Schaul, Matteo Hessel, Hado Van~Hasselt, Marc Lanctot, and Nando
  De~Freitas.
\newblock Dueling network architectures for deep reinforcement learning.
\newblock {\em International conference on machine learning (ICML)}, 2016.

\bibitem{zhao2019maximum}
Rui Zhao, Xudong Sun, and Volker Tresp.
\newblock Maximum entropy-regularized multi-goal reinforcement learning.
\newblock In {\em International Conference on Machine Learning}, pages
  7553--7562. PMLR, 2019.

\bibitem{zhao2018energy}
Rui Zhao and Volker Tresp.
\newblock Energy-based hindsight experience prioritization.
\newblock {\em Conference on Robot Learning (CoRL)}, 2018.

\end{thebibliography}
	
\end{document}